\documentclass[journal]{IEEEtran}

\usepackage{bm}
\usepackage{amsmath}
\usepackage{amsfonts}

\usepackage{url}
\usepackage{hyperref}
\usepackage{cite}

\usepackage{algorithmic}
\usepackage{amssymb}

\usepackage{graphicx}
\usepackage{float}
\usepackage{array}

\usepackage{booktabs}
\usepackage{multirow}
\usepackage{subcaption}
\usepackage{xcolor}

\usepackage{xspace}
\makeatletter
\DeclareRobustCommand\onedot{\futurelet\@let@token\@onedot}
\def\@onedot{\ifx\@let@token.\else.\null\fi\xspace}

\makeatother

\begin{document}
\title{DRAN: Detailed Region-Adaptive Normalization for Conditional Image Synthesis}
\author{Yueming Lyu,
        Peibin Chen, 
		Jingna Sun,
        Bo Peng,~\IEEEmembership{Member,~IEEE},\\
        Xu Wang,
        and Jing Dong,~\IEEEmembership{Senior Member,~IEEE}
\thanks{Yueming Lyu, Bo Peng, Jing Dong~(corresponding author) are with the Center for Research on Intelligent Perception and Computing~(CRIPAC), State Key Laboratory of Multimodal Artificial Intelligence Systems, Institute of Automation, Chinese Academy of Sciences~(CASIA), Beijing 100190, China, and also with the School of Artificial Intelligence, University of Chinese Academy of Sciences, Beijing 100049, China (E-mail: yueming.lv@cripac.ia.ac.cn; bo.peng@nlpr.ia.ac.cn; jdong@nlpr.ia.ac.cn).

Peibin Chen, Jingna Sun and Xu Wang are with ByteDance Inc, Beijing 100191, China (E-mail: chenpeibin@bytedance.com; 850248724@qq.com; wangxu.ailab@bytedance.com).
}
}


\maketitle

\begin{abstract}
	In recent years, conditional image synthesis has attracted growing attention due to its controllability in the image generation process. Although recent works have achieved realistic results, most of them have difficulty handling fine-grained styles with subtle details. To address this problem, a novel normalization module, named Detailed Region-Adaptive Normalization~(DRAN), is proposed. It adaptively learns both fine-grained and coarse-grained style representations. Specifically, we first introduce a multi-level structure, Spatiality-aware Pyramid Pooling, to guide the model to learn coarse-to-fine features. Then, to adaptively fuse different levels of styles, we propose Dynamic Gating, making it possible to adaptively fuse different levels of styles according to different spatial regions. Finally, we collect a new makeup dataset (Makeup-Complex dataset) that contains a wide range of complex makeup styles with diverse poses and expressions. To evaluate the effectiveness and show the general use of our method, we conduct a set of experiments on makeup transfer and semantic image synthesis. Quantitative and qualitative experiments show that equipped with DRAN, simple baseline models are able to achieve promising improvements in complex style transfer and detailed texture synthesis. Both the code and the proposed dataset will be available at \url{https://github.com/Yueming6568/DRAN-makeup.git}.
\end{abstract}


\IEEEpeerreviewmaketitle

\begin{figure}[htb]
	\centering
	  \includegraphics[width=1\columnwidth]{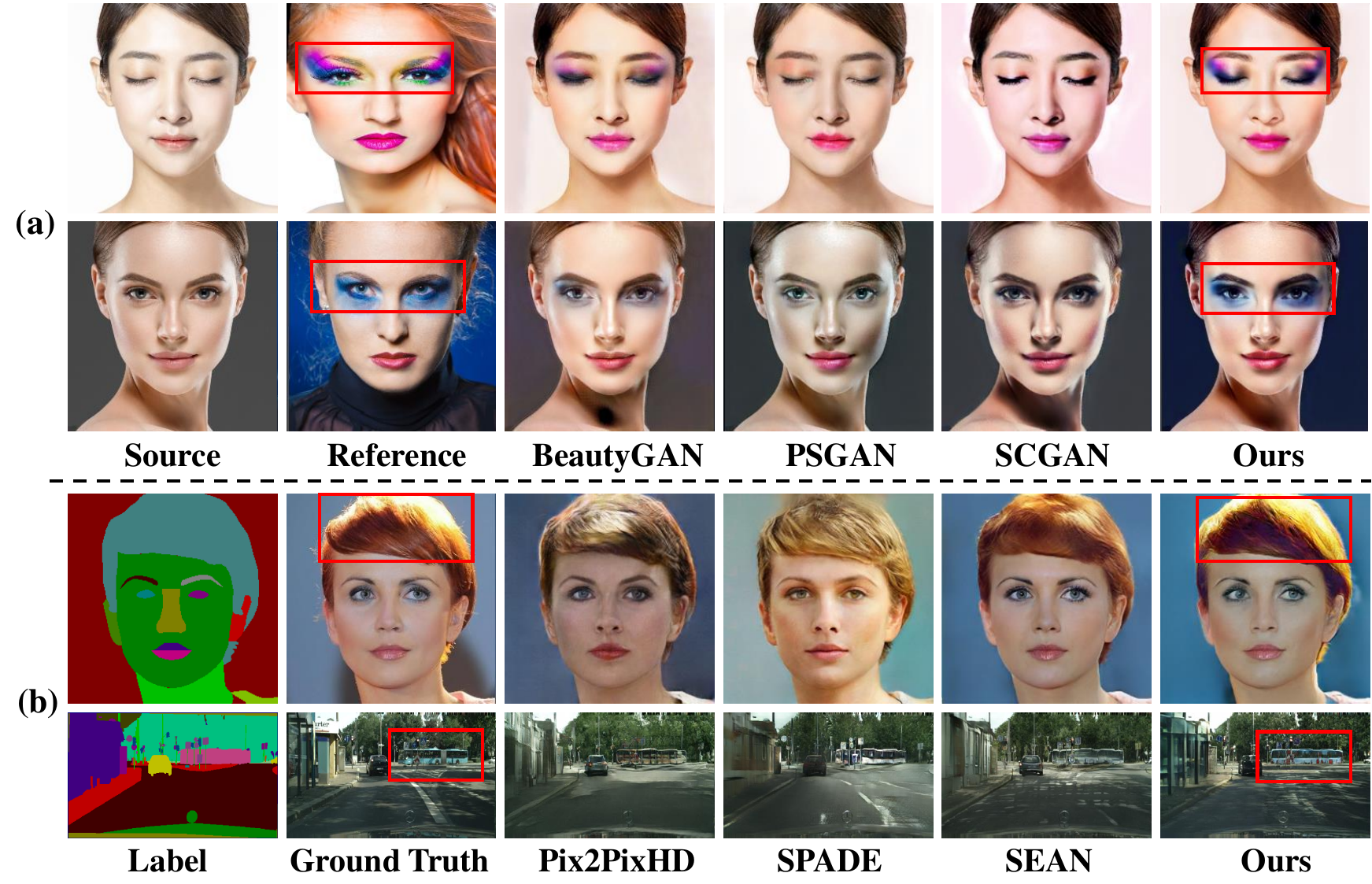}
	  \caption{Comparisons with state-of-the-art methods on two conditional image synthesis tasks, (a) makeup transfer and (b) semantic image synthesis. The results show that our method yields the most impressive results on complex styles~(a) and details synthesis~(b), such as heavy eye makeups~(a), the texture and reflection of hairs~(b), and the texture of the bus~(b), which are labeled with red boxes. Note that all the comparison methods in semantic image synthesis use the same style image~(the same one with the ground-truth image) as the style input. }
	\label{fig:shouye}
  \end{figure}

\section{Introduction}

\IEEEPARstart{I}{mage} synthesis has recently made great progress due to the rapid development of generative adversarial networks (GANs)~\cite{goodfellow2014generative}. 
Lots of variants of unconditional GANs are able to generate photo-realistic images from random latent codes~\cite{karras2017progressive,brock2018large,karras2019style}.
On the other hand, conditional image synthesis, aiming at generating images given conditional inputs, \emph{e.g.}, semantic labels~\cite{choi2018stargan,he2019attgan,chen2021semantic,liu2022isf}, guidance texts~\cite{zhang2017stackgan,hong2018inferring, yuan2019ckd, li2020exploring}, and reference images~\cite{wang2018high,park2019semantic,huang2020semantic,deng2022semantic}. 
Our work focuses on conditional image synthesis with reference images as conditions, which is still challenging. 
For example, in makeup transfer~\cite{li2018beautygan,chen2019beautyglow,gu2019ladn,lyu2021sogan,deng2021scgan}, the reference makeup should be accurately and effectively transferred to the source face. 
In semantic image synthesis~\cite{wang2018high,park2019semantic,zhu2020sean,zhu2020semantically,tan2021diverse,wang2021image}, the generated texture in each semantic region need to be precisely controlled by the reference style image. 

One of the key problems is \emph{how to effectively encode styles from the conditional reference images}. 
A well-designed style encoding mechanism contributes to fine-grained control and vivid image generation. 
Starting from AdaIN~\cite{huang2017arbitrary}, numerous existing approaches~\cite{brock2018large,huang2018multimodal,karras2019style,wang2018recovering,zhang2019self,park2019semantic,zhu2020sean,deng2021scgan} mainly consider encoding the styles into instance normalization parameters, which have been found to carry style information of images~\cite{gatys2016image,li2016combining, li2017demystifying}.
Some methods~\cite{huang2017arbitrary,deng2021scgan,CelebAMask-HQ,huang2018multimodal} encode the styles into channel-wise normalization parameters in the whole spatial space. 
However, this results in the spatial information being smoothed and many details being lost. 
For example, SCGAN~\cite{deng2021scgan}, an AdaIN~\cite{huang2017arbitrary}-based makeup transfer model, has difficulty in transferring heavy eye makeup (shown in Fig.~\ref{fig:shouye} (a)). 
To encode styles more effectively, some methods~\cite{zhu2020sean,men2020controllable,ling2021region,lv2021learning, yu2020region} compute the normalization parameters in region-wise spatial spaces instead of the whole spatial space. 
However, the parameters in the same region are also the same, and the encoded styles are not fine-grained enough.
On the other hand, in these methods, different semantic regions are treated equally, leading to regions with rich details being learned insufficiently.
For example, SEAN~\cite{zhu2020sean}, a region-adaptive normalization in semantic image synthesis, has difficulty in transferring the reflection of the hairs and the texture of the bus (shown in Fig.~\ref{fig:shouye} (b)).

In order to tackle the problem of effective and fine-grained style encoding, we propose a novel normalization module, called  Detailed Region-Adaptive Normalization (DRAN).
Different from the methods mentioned above, DRAN is designed to adaptively perform coarse-to-fine style encoding.
It has the ability to transfer fine-grained styles (\emph{e.g.} multiple makeup colors in a small region, hair textures and reflections) as well as coarse-grained styles (\emph{e.g.} general color), bringing an impressive improvement to the quality of the generated images, as shown in Fig.~\ref{fig:shouye}. 
Specifically, DRAN consists of Spatiality-aware Pyramid Pooling (SAPP) and Dynamic Gating.
SAPP first builds a style pyramid to represent multiple levels of styles. 
It calculates style parameters in different spatial pyramid levels instead of a single spatial level.
In this way, the coarse pyramid level captures the general styles and the fine pyramid level captures the fine-grained details. 
Then, to fuse these styles, a straightforward choice is to average them. 
But this solution is not the best choice, since the optimal fusion of style levels varies from region to region, and from case to case.
Therefore, we introduce a dynamic mechanism, Dynamic Gating, to dynamically ensemble multiple levels of style.
It learns which level is more important for each region of the current reference input, and fuses the features accordingly.
Finally, the extracted coarse-to-fine style representation flows to the next modules of the conditional synthesis model and contributes to final high-quality results. 

In summary, the main contributions are as follows:
\begin{itemize}
	\item We propose a simple yet effective module, called Detailed Region-Adaptive Normalization (DRAN), to adaptively handle different levels of styles, from coarse to fine. Besides, DRAN is a plug-and-play module for existing conditional image synthesis networks.
	\item In order to better evaluate DRAN on makeup transfer, we contribute a new makeup dataset (Makeup-Complex dataset, MC dataset for short), which contains a wide range of complex makeup styles with different poses and expressions. 
	\item Experiments show that equipped with DRAN, simple baseline models are able to generate more high-quality images. In makeup transfer, DRAN reduces the Proportionate Face Distance Metric (PFDM) of the baseline from 0.093 to 0.069 on the MC dataset. In semantic image synthesis, it reduces the Fréchet Inception Distance (FID) of the baseline from 20.45 to 15.84 on the CelebAMask-HQ dataset and from 52.88 to 48.92 on the CityScapes dataset. 
\end{itemize}

\section{Related Work}
\subsection{Conditional Image Synthesis}
Unlike image generation from random latent codes, conditional image synthesis aims at generating images based on the conditional input such as labels~\cite{choi2018stargan,he2019attgan,chen2021semantic,liu2022isf}, texts~\cite{zhang2017stackgan,hong2018inferring, yuan2019ckd, li2020exploring} and images~\cite{wang2018high,park2019semantic,huang2020semantic,deng2022semantic}.
Our work focuses on conditional image synthesis with images as conditions, \emph{e.g.}, makeup transfer and semantic image synthesis.

\vspace{1ex}
\textbf{Makeup transfer} aims to extract the makeup style from the reference image and transferring it to the source image. 
Since there are no paired data, many current methods apply the architecture of CycleGAN~\cite{zhu2017unpaired} to transfer makeup styles in an unsupervised manner.
In particular, BeautyGAN~\cite{li2018beautygan} introduces a dual input/output framework to achieve makeup transfer and makeup removal simultaneously and a pixel-level histogram loss on facial components to refine local details.
PairedCycleGAN~\cite{chang2018pairedcyclegan} warps the reference image to the source image to produce pseudo transferred images for guiding makeup transfer.
BeautyGlow~\cite{chen2019beautyglow} leverages the Glow~\cite{kingma2018glow} architecture to decompose makeup and non-makeup latent vectors and then invert the recombined latent vectors to the image domain. 
LADN~\cite{gu2019ladn} incorporates multiple overlapping local discriminators and asymmetric losses for dramatic makeup transfer.
However, the above approaches simply fuse the extracted makeup styles to the source domain by concatenation and thus inevitably lead to the spatial misalignment problem. 
In order to transfer between images under different poses and expressions, PSGAN~\cite{jiang2020psgan} employs an Attentive Makeup Morphing~(AMM) module to align the makeup styles with the source features and produces spatial-adaptive modulation parameters with style statistics.
However, the proposed AMM applies pixel-to-region alignment, which is ambiguous and cannot produce accurately aligned parameters for makeup transfer, which is also found in \cite{deng2021scgan}.
SOGAN~\cite{lyu2021sogan} proposes a 3D-aware approach for shadow and occlusion robust makeup transfer.  
SCGAN~\cite{deng2021scgan} achieves spatially-invariant makeup transfer by encoding component-wise style information into multiple AdaIN parameters that discard the spatial information.
\begin{figure*}[t]
    \centering
		\includegraphics[width=1\textwidth]{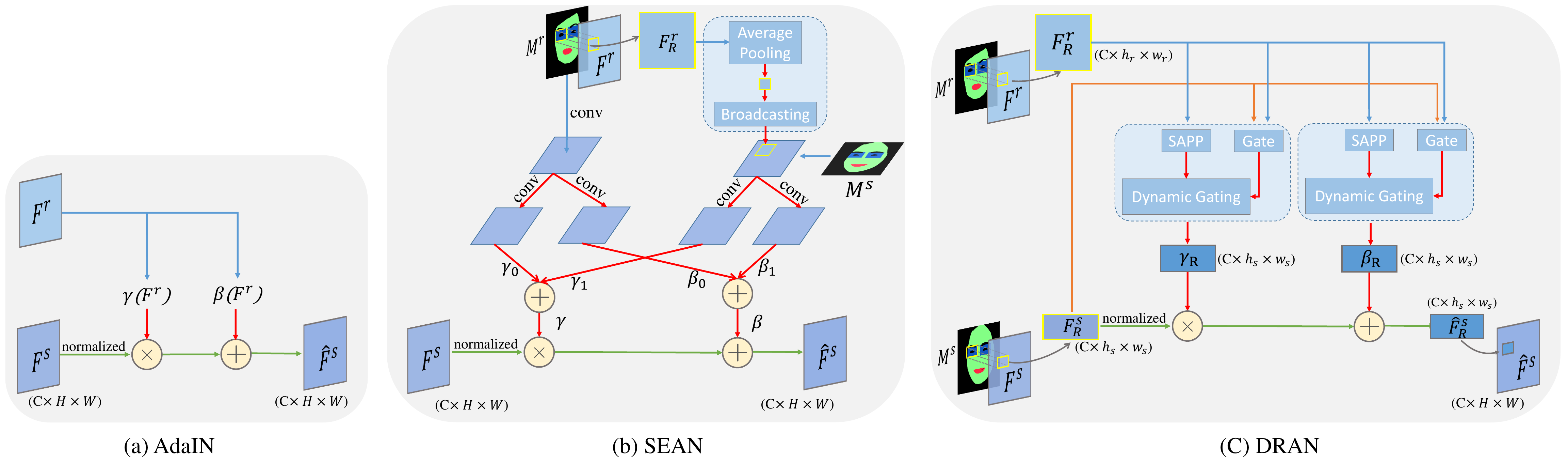}
	\caption{The structures of Adaptive Instance Normalization~(AdaIN)~\cite{huang2017arbitrary}~(a), Semantic Region-adaptive Normalization~(SEAN)~\cite{zhu2020sean}~(b), and our proposed Detailed Region-adaptive Normalization~(DRAN)~(c).
	Compared with (a) and (b), our method first applies semantic segmentation masks to obtain corresponding regional features for both source and reference features. Then, by applying Spatiality-aware Pyramid Pooling~(SAPP) and Dynamic Gating, multi-level modulation parameters are extracted from the reference features and dynamically fused to modulate the source features. More details of DRAN are described in Sec.~\ref{sec}.}
	\label{fig:pipeline_all}
\end{figure*}

\vspace{1ex}
\textbf{Semantic image synthesis} focuses on converting semantic segmentation maps to photo-realistic images. 
One notable work is Pix2Pix~\cite{isola2017image}, which adopts an encoder-decoder generator that takes semantic masks as input, and utilizes a PatchGAN discriminator.  
Pix2PixHD~\cite{wang2018high}, as an extended version of Pix2Pix, proposes coarse-to-fine generators and discriminators to produce high-resolution images from semantic masks, where the reference style image can be encoded by an encoder network and transferred to the semantic mask information by instance-wise average pooling. 
SPADE~\cite{park2019semantic} proposes a spatially-adaptive normalization method that modulates the activations using input semantic masks through a spatially-adaptive learned transformation.
By attaching an encoder, the reference style image can be processed into a vector, fed to the generator, and combined with the segmentation mask information via the proposed normalization. 
However, SPADE uses only one style code to control the whole modulation parameters, which is not sufficient for detailed per-region control. 
Therefore, SEAN~\cite{zhu2020sean}, an improvement of SPADE, proposes a semantic region-adaptive normalization to enable control of per region semantic information. 
It first broadcasts the styles of each region according to the semantic mask and then creates per pixel normalization similar to SPADE. 

Though most of the above methods have achieved promising results, their results still lack accurate details. For instance, as shown in Fig.~\ref{fig:shouye}, SCGAN has difficulty in transferring complex eye makeup, and SEAN is hard to transfer the reflection of hairs.
Based on SEAN, we propose a novel conditional normalization to enable the transfer of more fine-grained styles within each region. 

\subsection{Normalization Layers}
In the deep learning era, normalization layers play a vital role in achieving better convergence and performance. They can be divided into unconditional normalizations and conditional normalizations. 
Typical unconditional normalization methods include Batch Normalization (BN)~\cite{ioffe2015batch}, Instance Normalization (IN)~\cite{ulyanov2016instance}, Layer Normalization (LN)~\cite{ba2016layer} and Group Normalization (GN)~\cite{wu2018group}, \emph{etc}. 
Different from unconditional normalization techniques, conditional normalizations require external input which is used to infer the modulation parameters. 
Due to its effectiveness in transferring feature statistics, it has been widely used in many conditional image synthesis tasks, such as style transfer~\cite{huang2017arbitrary}, makeup transfer~\cite{jiang2020psgan, deng2021scgan}, semantic image synthesis~\cite{park2019semantic, zhu2020sean, tan2021diverse}. 
As a seminal work, AdaIN~\cite{huang2017arbitrary} is proposed to encode the conditional input into channel-wise global modulation parameters. 
After that, SPADE proposes the first spatially-adaptive instance normalization to learn style parameters with spatial information for semantic image synthesis. 
Inspired by SPADE, some variations~\cite{zhu2020semantically, zhu2020sean, tan2021efficient, tan2021diverse} are proposed.
For example, GroupDNet~\cite{zhu2020semantically} introduces conditional group normalization by replacing typical convolutions to group convolutions~\cite{krizhevsky2012imagenet}. 
SEAN~\cite{zhu2020sean} proposes a region-adaptive instance normalization method. 
CLADE~\cite{tan2021efficient} proposes a class-adaptive instance normalization to be adaptive to different semantic classes.  
INADE~\cite{tan2021diverse} proposes a instance-adaptive normalization to achieve instance style control. 
Different from these methods, we integrate Spatiality-aware Pyramid Pooling and Dynamic Gating into our proposed normalization, which aims to learn coarse-to-fine style representation adaptively.
\begin{figure*}[t]
    \centering
		\includegraphics[width=0.9\textwidth]{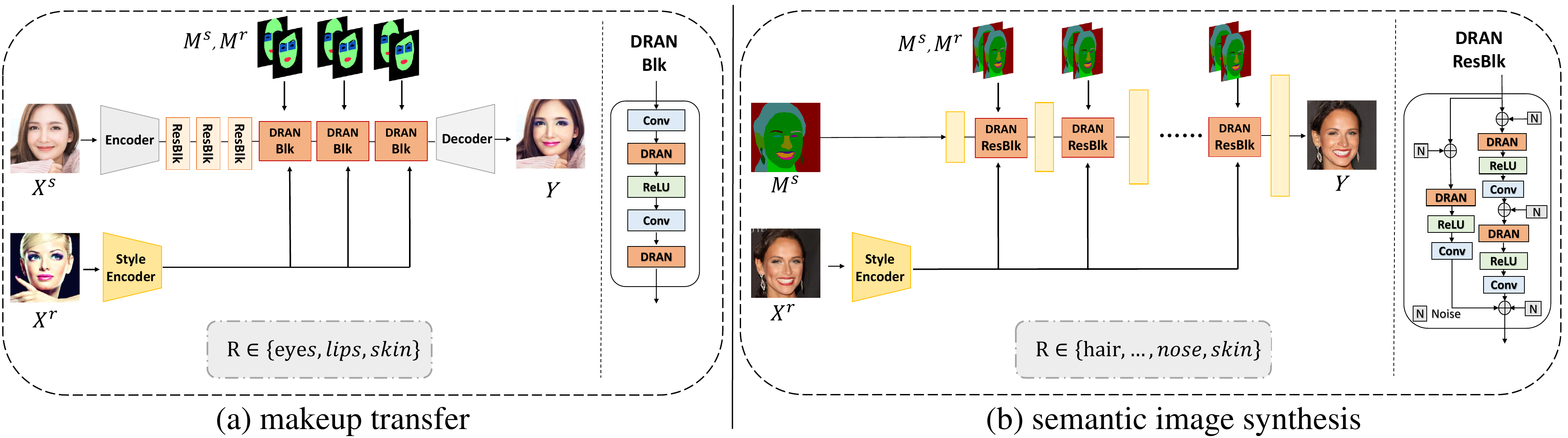}
	\caption{The overall network architectures equipped with DRANs on (a) makeup transfer and (b) semantic image synthesis. We only show the generator parts for simplicity.
	Specifically, the style encoder extracts style features from the conditional reference input $X^r$. 
	After that, DRANs contained in DRAN-blocks or DRAN-Resblocks perform region-adaptive style encoding guided by semantic maps of the source and reference input, $M^s$ and $M^r$.  
	For makeup transfer, the semantic regions are generally split into eyes, lips and skin. For semantic image synthesis, the semantic regions are hair, nose, skin, \emph{etc}. 
	Finally, the generated images are supervised by task-specific losses, such as makeup loss for makeup transfer and reconstruction loss for semantic image synthesis. 
	}
	\label{fig:framework}
\end{figure*}
\section{Preliminaries}
In this section, we provide an introduction of the normalization approaches for style encoding in conditional image synthesis, which mainly include adaptive instance normalization and region-adaptive instance normalization.
Moreover, we show the structures of two typical normalization methods and our proposed normalization method in Fig.~\ref{fig:pipeline_all}. 

\textbf{Adaptive Instance Normalization~(AdaIN).}
The structure of AdaIN is depicted in Fig.~\ref{fig:pipeline_all}~(a). As a classical method for style encoding, it was first proposed for real-time style transfer task~\cite{huang2017arbitrary} and later widely adopted in various vison tasks~\cite{brock2018large,huang2018multimodal,karras2019style,wang2018recovering,zhang2019self}. 
Let $F^s$ and $F^r$ denote the $i$-th layer activations in a deep convolutional network of the source input and the reference input, respectively.
For simplicity, the symbol of $i$ is omitted.
Let $C$, $H$ and $W$ be the number of channels, height, and width of $F^{s}$, respectively. 
The modulated activation value $\hat{F}_{c, h, w}^{s}$ at the position $(c, h, w)$ can be computed as
\begin{equation}
	\hat{F}_{c, h, w}^{s}=\gamma_{c}(F^r_c) \frac{F_{c, h, w}^{s}-\mu_{c}}{\sigma_{c}}+\beta_{c}(F^r_c),
\end{equation}
where $\gamma_{c}(F^r_c)$ and $\beta_{c}(F^r_c)$ are produced modulation parameters from $F^r_c$ in channel $c$. 
$\mu_{c}$ and $\sigma_{c}$ are the mean and standard deviation of the activation $F_{c, h, w}^s$ in channel $c$:
\begin{equation}
	\begin{aligned}
	\mu_{c} &=\frac{1}{HW} \sum_{h,w} F_{c,h,w}^s, \\
	\sigma_{c}&=\sqrt{\frac{1}{HW} \sum_{h,w}\left(F_{c,h,w}^s-\mu_{c}\right)^{2}+\epsilon}.
	\end{aligned}
\end{equation}
Note that modulation parameters $\gamma(F^r)$ and $\beta(F^r)$ and normalization parameters $\mu$ and $\sigma$ are all applied in a spatially invariant way to activations at all spatial coordinates.
Therefore, AdaIN-based methods neglect spatial information and only encode global style information.

\textbf{Region-Adaptive Instance Normalization.}
Recently, Region-adaptive Instance Normalization~\cite{zhu2020sean,lv2021learning,ling2021region} has been proposed to encode and control the style of each region individually.
It calculates the modulated activation $\hat{F}_{c,h,w}^{s}(\textit{M})$ at site $(c, h, w)$ in the semantic region $\textit{M}$ according to regional information:
\begin{equation}
	\hat{F}_{c,h,w}^{s}(\textit{M})=\gamma_{c,h,w}(\textit{M}) \frac{F_{c,h,w}^s(\textit{M})-\mu_{c}}{\sigma_{c}}+\beta_{c,h,w}(\textit{M}).
\end{equation}
where $\gamma_{c,h,w}(\textit{M})$ and $\beta_{c,h,w}(\textit{M})$ are region-based modulation parameters of $\textit{M}$.
Still, the modulation parameters computed by Region-adaptively Instance Normalization are usually the same in the same region, and thus the encoded styles are not fine-grained enough. 
Take SEAN~\cite{zhu2020sean} as an example, as shown in Fig.~\ref{fig:pipeline_all}~(b), it first extracts regional features $F_R^r$ according to mask $M^r$ and then reduces the features to global regional features by a region-wise average pooling layer.
After that, the global regional features are broadcasted to the corresponding regions according to $M^s$ and yield modulation parameters by convolution layers. 
However, since the reference regional features are pooled in each region, the produced modulation parameters cannot represent more detailed styles. 

To encode and control more detailed styles from conditional inputs, we propose a novel normalization module, called Detailed Region-Adaptive Normalization (DRAN). 
As shown in Fig.~\ref{fig:pipeline_all}~(c), compared with AdaIN, DRAN considers per-region modulation parameters rather than global ones. 
Compared with SEAN, DRAN encodes coarse-to-fine styles in each region and fuses multi-level styles adaptively for different semantic regions. More details of the proposed DRAN will be described in the following section. 

\section{Methodology}
\label{sec}
Our goal is to adaptively learn coarse-to-fine style representation for conditional image synthesis. To achieve this goal, we design a novel normalization module, called Detailed Region-Adaptive Normalization (\ref{sec1}), which is composed of Spatiality-aware Pyramid Pooling (\ref{sec2}) and Dynamic Gating (\ref{sec3}).
Specifically, we apply DRAN onto two baseline models, SCGAN~\cite{deng2021scgan} and SEAN~\cite{zhu2020sean}, for makeup transfer and semantic image synthesis, respectively. 
Details about the models are described in the captions of Fig.~\ref{fig:framework}. 
We follow the training strategy and network structure of two baseline models.
The task-specific losses are described in Sec.~\ref{sec5}.

\subsection{Detailed Region-adaptive Normalization}
\label{sec1}

As depicted in Fig.~\ref{fig:pipeline_all}, $F^s \in \mathbb{R}^{C \times H \times W}$ and $F^r \in \mathbb{R}^{C \times H \times W}$ are the $i$-th layer features of the source $X^s$ and the reference input $X^r$ respectively. 
For simplicity, the symbol of $i$ is omitted.
Let $C$, $H$ and $W$ be the number of channels, height and width of the feature map. $M^s \in \mathbb{R}^{1 \times H \times W}$ and $M^r\in \mathbb{R}^{1 \times H \times W}$ are their segmentation maps resized to the same sizes as the feature maps.
Firstly, $F^s$ and $F^r$ are cropped to obtain regional features $F_{R}^s\in \mathbb{R}^{C \times h_r \times w_r}$ and $F_{R}^r\in \mathbb{R}^{C \times h_s \times w_s}$ of each semantic region according to $M^s$ and $M^r$, \emph{e.g.}, the region of eyes. 
Then, take $F_R^r$ as input, Spatiality-aware Pyramid Pooling (SAPP) is proposed to capture different granularity of style information from the reference input. 
After that, a lightweight module, called Dynamic Gating, is introduced to adaptively fuse the outputs of SAPP to obtain the modulation parameters $\gamma_R$ and $\beta_R$. 
Then, the new activation features of region R, denoted by $\hat{F}_R^s$, can be obtained by
\begin{align}\label{eq:dran1}
    &\hat{F}_R^s=\gamma_R \frac{F_R^s-\mu_R}{\sigma_R}+\beta_R,
\end{align}
where $\mu_R$ and $\sigma_R$ are region-wise mean and variance parameters of the source features $F_R^s$. They can be computed by
\begin{equation}
	\begin{aligned}
	\mu_{R} &=\frac{1}{N_R} \sum_{h,w \in R} F_{c,h,w}^s, \\
	\sigma_{R}&=\sqrt{\frac{1}{N_R} \sum_{h,w \in R}\left(F_{c,h,w}^s-\mu_{R}\right)^{2}+\epsilon}.
	\end{aligned}
	\text{,}
\end{equation}
where $N_R$ denotes the number of pixels in region $R$ of the source features $F^s_R$ and $\epsilon$ is a very small constant.
Finally, the new activation features $\hat{F}^s$ is obtained by compositing all modulated regional features according to the semantic mask $M^s$.

Since $\gamma_R$ and $\beta_R$ are produced similarly, without loss of generality, we only take $\beta_R$ as the example and illustrate the details of Spatiality-aware Pyramid Pooling and Dynamic Gating in Fig.~\ref{fig:DRAN}.

\subsection{Spatiality-aware Pyramid Pooling}
\label{sec2}
\begin{figure}[tp]
	\centering
		\includegraphics[width=0.85\columnwidth]{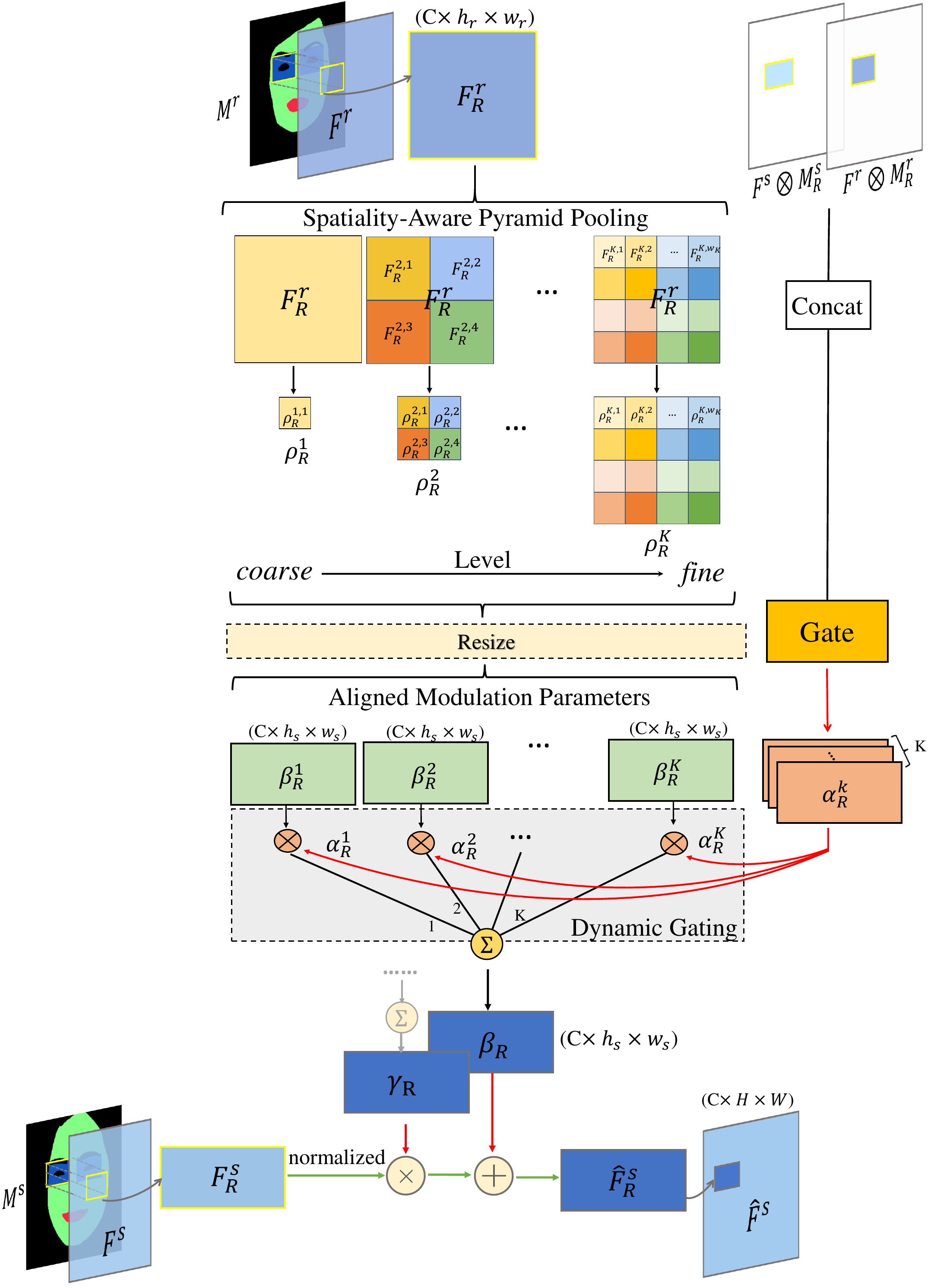}
		\caption{Detailed illustration of our proposed DRAN. Take region-wise features as input, SAPP first learns coarse-to-fine modulation parameters$\rho^k_{R}$ by building a spatial pyramid for the input features with $K$ branches. Then, the aligned parameters $\beta_R^k$ are obtained by resizing $\rho^k_{R}$ to the size of $F_R^s$. Finally, Dynamic Gating fuses the aligned modulation parameters of different branches in an adaptive way to output the final modulation parameter $\beta_R$.}
	\label{fig:DRAN}
\end{figure}

Current methods for extracting channel-wise normalization parameters discard the spatial information within a particular region, thus avoiding the spatial misalignment problem between corresponding source and reference regions. 
Instead, we learn subtle details by considering multi-level spatial information within a region. 
Under this situation, the different spatial locations and the different region sizes of corresponding regions may cause misalignment. Specifically, take region $R$ in Fig.~\ref{fig:DRAN} as an example, the source region and the reference region in their original images have different spatial locations. Therefore, directly transferring using the whole image features may cause some artifacts in the resulting image since they are not spatially aligned (such as BeautyGAN~\cite{li2018beautygan} and LADN~\cite{gu2019ladn}). In addition, the regional features $F_R^s$ and $F_R^r$ often have different spatial sizes (namely ($h_s$, $w_s$) are not equal to ($h_r$, $w_r$)), thus transferring modulation parameters between different sizes may also have the spatial misalignment problem.
In this section, we introduce Spatiality-aware Pyramid Pooling (SAPP), which extracts coarse-to-fine style representation while alleviating the misalignment problem.

As has been described in \ref{sec1}, region-wise features, \emph{e.g.} $F_R^s$, are cropped from the whole image feature, \emph{e.g.} $F^s$. 
In this way, the regional features $F_R^s$ and $F_R^r$ are coarsely aligned in the region-wise sense, which avoids the misalignment caused by different spatial locations between corresponding regions.
After that, inspired by SPP-net~\cite{he2015spatial}, the corresponding regional features with different sizes are pooled to multiple fixed-size spatial levels in the same way.
Then, the modulation parameters are computed in each spatial pyramid level, thus the misalignment caused by different region sizes is alleviated.
Finally, the coarse-to-fine style representations, \emph{i.e.} the modulation parameters, are extracted as the mean and variance of each feature block and then resized to the spatial size of the source regional feature map.
Experiments show that our method achieves makeup transfer accurately even for input images with large spatial misalignment (shown in Fig.~\ref{fig:comparsion}).

In detail, SAPP contains $K$ branches that produce modulation parameters at various levels independently. 
Take the branch $k$ as an example, the input feature $F_R^r$ is partitioned into $N_B = w_k \times w_k$ blocks.
$F_R^{k,b}\in \mathbb{R}^{C\times\frac{h_r}{w_k} \times \frac{w_r}{w_k}}$ is the $b^{th}$ block of the $k^{th}$ branch.
Let $N_P=\frac{h_r}{w_k} \times \frac{w_r}{w_k}$ be the number of pixels in $F_R^{k,b}$, modulation parameters $\rho^{k}_{R}\in \mathbb{R}^{C\times w_k \times w_k}$ (shown as the example in Fig.~\ref{fig:DRAN}) and $\tau^{k}_{R}\in \mathbb{R}^{C\times w_k \times w_k}$ (omitted in Fig.~\ref{fig:DRAN} for simplicity) are computed via the following equation:
\begin{align}\label{eq:sapp_beta}
&\rho^{k,b}_{R} = \frac{1}{N_P} \sum_{p} F_R^{k,b,p},\\
&\tau^{k,b}_{R} = \sqrt{\frac{1}{N_{P}} \sum_{p} (F_R^{k,b,p} - \rho^{k,b}_{R})^2 + \epsilon},
\end{align}
where $\rho^{k,b}_{R}\in \mathbb{R}^{C\times 1\times 1}$ and $\tau^{k,b}_{R}\in \mathbb{R}^{C\times 1\times 1}$ are the $b^{th}$ value of $\rho^{k}_{R}$ and $\tau^{k}_{R}$  respectively, and $\epsilon$ is a very small constant. 
Finally, the modulation parameters $\rho^{k}_{R}$ and $\tau^{k}_{R}$ are resized to the size of $F_{R}^s$ to obtain the aligned parameters $\beta^k_R$ and $\gamma^k_R$. It is noted that we use the nearest interpolation for resizing the modulation parameters in our implementation, and according to the following experiments, different interpolation methods make no significant differences.

\subsection{Dynamic Gating}
\label{sec3}
By far, the outputs of SAPP contain modulation parameters extracted from different pyramid levels. Directly averaging them over levels to obtain the final modulation parameters is not the best choice, for the optimal combination of style levels varying from region to region, and from case to case.
Take makeup transfer as an example, the eyes region usually has more fine-grained styles (\emph{e.g.} multiple colors in eye shadows) than the lips region which usually has a uniform color. 
Moreover, for the eyes region, some reference eye shadows contain rich details and multiple colors, and some reference eye shadows only have simple styles.  
Therefore, to improve the adaptiveness of fusing multiple levels of style parameters, we introduce a dynamic gating mechanism to adaptively ensemble multiple style levels for each semantic region according to its complexity of details.

Specifically, the Gate is a lightweight network composed of convolution and softmax operations. 
We first obtain the regional features from the masking of the original features, which can be obtained by $F^s \otimes M_R^s$ and $F^r \otimes M_R^r$, where $M_R^s \in \mathbb{R}^{1 \times H \times W}$ and $M_R^r \in \mathbb{R}^{1 \times H \times W}$ are their segmentation masks for region $R$. Then, take the concatenation of them as inputs, the Gate calculates the weights of the branches as follows:
\begin{align}
    \alpha_R^k = \text{Softmax}(\text{Conv}_\theta(\text{Cat}(F^s \otimes M_R^s, F^r \otimes M_R^r))),\\
    w_R^k = \text{Softmax}(\text{Conv}_\theta(\text{Cat}(F^s \otimes M_R^s, F^r \otimes M_R^r))),
\end{align}
where the values of $\alpha_R^k$ and $w_R^k$ $\in [0,1]$.
Then, the outputs of the $K$ branches from SAPP are combined to obtain the final modulation parameters, 
\begin{equation}
\label{equ:dg}
	\begin{aligned}
    &\beta_R =\sum_{k=1}^{K} \alpha_R^k \beta_R^k \text{,}\quad \\
	&\gamma_R =\sum_{k=1}^{K} w_R^k \gamma_R^k\text{,}
	\end{aligned}
\end{equation}
where $\alpha_R^k$ and $w_R^k$ have the same spatial size as $\beta_R^k$ and $\gamma_R^k$. 

\subsection{Loss Functions}
\label{sec5}
In the following, we introduce the loss functions used in the two tasks.

\textbf{For makeup transfer},
let $X^s$ and $X^r$ denote the source image and the reference makeup image, respectively.
$Y=G(X^{s}, X^{r})$ is the transferred result with the makeup style of $X^{r}$ and the facial identity of $X^{s}$.
In this expression, the input semantic masks are omitted for simplicity.

\textit{Cycle Consistency Loss.}
Generally, there are no paired data for supervising makeup transfer.
We adopt the cycle consistency loss~\cite{zhu2017unpaired} to guide the network in an unsupervised way. 
It is formulated as
\begin{equation}
    \begin{aligned}
    \mathcal{L}_{cyc} &=\|G(G(X^{s}, X^{r}), X^{s})-X^{s}\|_{1} \\
    &+\|G(G(X^{r}, X^{s}), X^{r})-X^{r}\|_{1} \text{,}
    \end{aligned}
  \end{equation}
where $\|\cdot\|_{1}$ represents the L1-Norm.

\textit{Makeup Loss.}
To add supervision during the makeup process, 
we utilize local Histogram Matching $HM(X^s \otimes M_R^s, X^r \otimes M_R^r)$ \cite{li2018beautygan} that provides a pseudo ground truth for the transferred image, 
where $M_R^s$ and $M_R^r$ are the segmentation masks for region $R$ in $X^s$ and $X^r$ respectively. 
The overall makeup loss consists of local $HM$ on three regions, \emph{i.e.}, eyes, lips, and skin:
\begin{small}
\begin{equation}
    \begin{aligned}
        \mathcal{L}_{makeup}&=\\
        &\sum_{R}\left\|(G_{R}(X^{s}, X^{r})-HM(X^s \otimes M_R^s, X^r \otimes M_R^r)\right\|_{2} \\
        &+\left\|G_{R}(X^{r}, X^{s})-HM(X^r \otimes M_R^r, X^s \otimes M_R^s)\right\|_{2} \text{,}
    \end{aligned}
\end{equation}
\end{small}
where $\|\cdot\|_{2}$ represents the L2-Norm, and $G_{R}$ represents obtaining the region $R$ from the output of the transfer model.

\textit{Perceptual Loss.}
In order to preserve the facial identity of the transferred image, we adopt the perceptual loss~\cite{johnson2016perceptual} to measure the differences between the source image and the transferred image in the feature space. 
Here, a VGG-16 model pre-trained on ImageNet is used as the feature extractor. 
Let $F^{i}$ represent the output of $i^{th}$ layer of the feature extractor. 
The perceptual loss is formulated as
\begin{equation}
	\begin{aligned}
\mathcal{L}_{per}&=\left\|F^{i}(G(X^{s}, X^{r}))-F^{i}(X^{s})\right\|_{2} \\
&+\left\|F^{i}(G(X^{r}, X^{s}))-F^{i}(X^{r})\right\|_{2} \text{.}
	\end{aligned}
\end{equation}

\textit{Adversarial Loss.}
For producing more realistic results, we apply the adversarial loss.
Here, two domain discriminators $D_{s}$ and $D_{r}$ are introduced for the non-makeup image domain and the makeup image domain, respectively.
They are formulated as
\begin{equation}
    \begin{aligned}
    \mathcal{L}_{adv}^{G}=&-\mathbb{E}\left[\log \left(D_{s}(G(X^{r}, X^{s}))\right)\right] \\
    &-\mathbb{E}\left[\log \left(D_{r}(G(X^{s}, X^{r}))\right)\right] \text{,} \\
    \mathcal{L}_{adv}^{D}=&-\mathbb{E}\left[\log D_{s}(X^{s})\right]-\mathbb{E}\left[\log D_{r}(X^{r})\right] \\
    &-\mathbb{E}\left[\log \left(1-D_{s}(G(X^{r}, X^{s}))\right)\right] \\
    &-\mathbb{E}\left[\log \left(1-D_{r}(G(X^{s}, X^{r}))\right)\right] \text{.}
    \end{aligned}
\end{equation}

\textit{Total Loss for Makeup Transfer.}
The total loss for makeup transfer can be expressed as:
\begin{equation}
    \begin{aligned}
    \mathcal{L}_{G} &= \lambda_{a}^m\mathcal{L}_{adv}^{G} + \lambda_{c}^m\mathcal{L}_{cyc} + \lambda_{m}^m\mathcal{L}_{makeup} + \lambda_{p}^m\mathcal{L}_{per} \text{,}\\
    \mathcal{L}_{D} &= \lambda_{a}^m\mathcal{L}_{adv}^{D} \text{,}
    \end{aligned}
    \label{con:totalmloss}
  \end{equation}
where $\lambda_{a}^m$, $\lambda_{c}^m$, $\lambda_{m}^m$, $\lambda_{p}^m$ are the trade-off parameters for makeup transfer.

\textbf{For semantic image synthesis}, let $X^r$ and $M^s$ denote the reference image and the source segmentation map.
$Y=G(M^s, X^r)$ means transferring the style of $X^r$ into the source semantic map $M^s$, and obtaining the output image $Y$.
During training, $M^s$ is the corresponding mask of $X^r$ and the training is formulated as a reconstruction problem.
During inference, $X^r$ and $M^s$ can be different.

\textit{Perceptual Loss.}
Perceptual loss is defined by the distance of the high-dimensional features between the real and synthesized images.
Here, a pretrained VGG-19 model on ImageNet is used as the feature extractor.
Let $F^{i}$ represent the output of $i$-th layer of the feature extractor. 
The perceptual loss is expressed as
\begin{equation}
	\begin{aligned}
\mathcal{L}_{per}&=\sum_{i=1}^T\left\|F^{i}(G(M^s, X^r))-F^{i}(X^r)\right\|_{1} \text{,}
	\end{aligned}
\end{equation}
where T is set to 5, which is the total number of layers used to calculate the perceptual loss.
\begin{figure*}[htb]
	\centering
	  \includegraphics[width=1\textwidth]{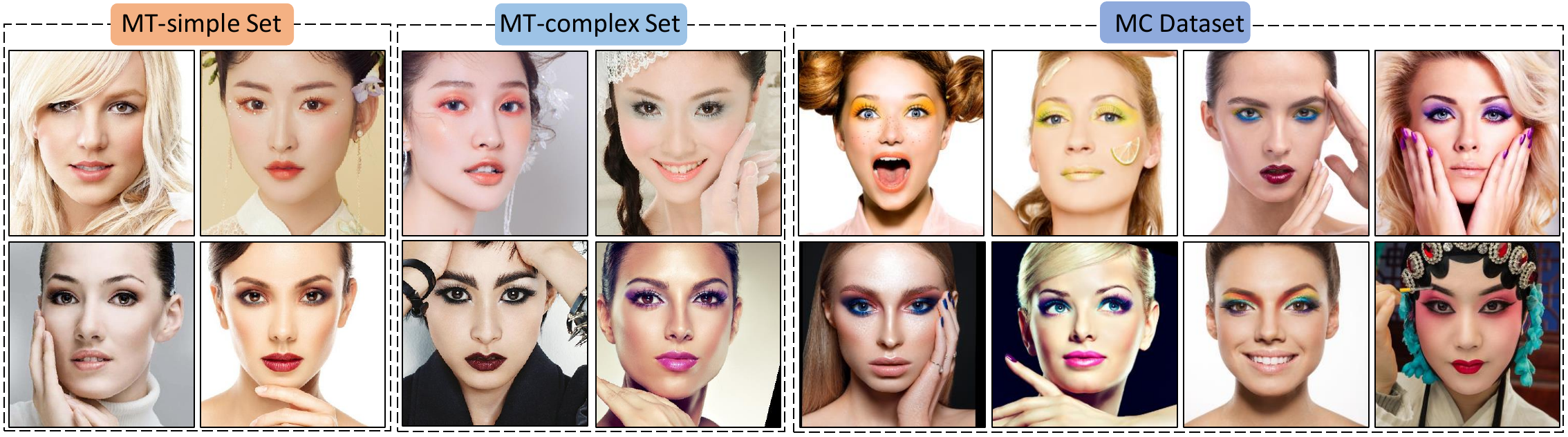}
	  \caption{Sampled makeup images from the MT dataset and the proposed MC dataset, where the MT dataset is split into subsets with simple makeup styles~(MT-simple set) and complex makeup styles~(MT-complex set).}
	\label{fig:data}
  \end{figure*}

\textit{Adversarial Loss.}
To guarantee the generation quality, we use two discriminators at different scales~\cite{wang2018high}, denoted as $D_k$, where $k=1,2$. 
The adversarial loss is built with the hinge loss and formulated as
\begin{equation}
  \begin{aligned}
  \mathcal{L}_{adv}^{G}=&-\mathbb{E}\left[\left(D_k(G(M^s, X^r))\right)\right] \text{,} \\
  \mathcal{L}_{adv}^{D}=&\mathbb{E}\left[max\left(0,1-D_k(M^s, X^r)\right)\right]+ \\
  &\mathbb{E}\left[max\left(0,1+D_k(G(M^s, X^r))\right)\right] \text{.}
  \end{aligned}
\end{equation}

\textit{Feature Matching Loss.}
Feature matching loss~\cite{wang2018high} is used for stablizing the training process. 
Let $D_k^i$ denote the $i$-th features of the discriminator $D_k$. 
It is formulated as 
\begin{equation}
  \begin{aligned}
    \mathcal{L}_{fm} = 
    &\sum_{k=1}^{2}\sum_{i=1}^{N}\frac{1}{N_{i}}\left\|(D_k^i(M^s,X^r)-D_k^i(M^s, G(M^s,X^r))\right\|_{1} \text{,}
  \end{aligned}
\end{equation}
$N$ is the total number of layers in the discriminator and $N_i$ is total number of pixels in each layer. 

\textit{Total Loss for Semantic Image Synthesis.}
The total loss for semantic image synthesis can be expressed as:
\begin{equation}
    \begin{aligned}
    \mathcal{L}_{G} &= \lambda_{a}^s\mathcal{L}_{adv}^{G} + \lambda_{fm}^s\mathcal{L}_{fm} + \lambda_{p}^s\mathcal{L}_{per} \text{,}\\
    \mathcal{L}_{D} &= \lambda_{a}^s\mathcal{L}_{adv}^{D} \text{,}
    \end{aligned}
    \label{con:totalsloss}
  \end{equation}
where $\lambda_{a}^s$, $\lambda_{fm}^s$, $\lambda_{p}^s$ are the trade-off parameters for semantic image synthesis.

\section{Experiments}
\subsection{Experiment Settings}
\textbf{Datasets.}
For makeup transfer, following previous methods~\cite{li2018beautygan, jiang2020psgan,deng2021scgan}, we also use MT (Makeup Transfer) dataset~\cite{li2018beautygan}, which contains 1,115 non-makeup images and 2,719 makeup images. 
We follow the train/test protocol of \cite{li2018beautygan}, randomly choosing 100 non-makeup images and 250 makeup images for testing.
The rest images are used for training and validation. 
During training, we randomly sample a pair of makeup image and non-makeup image from the training set to perform instance-wise makeup transfer.
Since the number of training pairs far exceeds the number of individual training images, the training can converge well on the MT dataset.

Furthermore, we let three professionals in makeup transfer to categorize the MT dataset with respect to the complex levels of makeups. 
As a result, there are about 2,330 images with simple makeup styles and only 389 (about 14\%) images have complex makeups with abundant details and multiple colors.
Because of this, the MT dataset is relatively easy and not very suitable for validating fine-grained makeup transfer effects of different methods.
Therefore, we collect a new dataset, Makeup-Complex (MC) dataset that contains a wide range of complex makeup styles under different poses and expressions. 
We collect 558 with-makeup images from the Internet, and then crop and resize the images to the size of $256 \times 256$.
It is an extra \emph{test set} for evaluating the performance of transferring fine-grained makeup styles. \emph{We will release it when the paper is published.}
Some examples of both datasets are visualized in Fig.~\ref{fig:data}. 

For semantic image synthesis, we conduct on the CelebAMask-HQ dataset~\cite{liu2015deep,karras2017progressive,CelebAMask-HQ} (with $256 \times 256$ resolution) and the CityScapes dataset~\cite{cordts2016cityscapes} (with $256 \times 512$ resolution).
CelebAMask-HQ contains 30,000 CelebAHQ face images and corresponding segmentation masks. There are 19 different region classes which include all facial components and accessories. 
CityScapes contains 3500 street scene images with 35 different region annotations.  
We follow the given train/test split for each dataset as defined in their papers.
\begin{figure*}[!tbp]
	\begin{center}
		\includegraphics[width=0.92\textwidth]{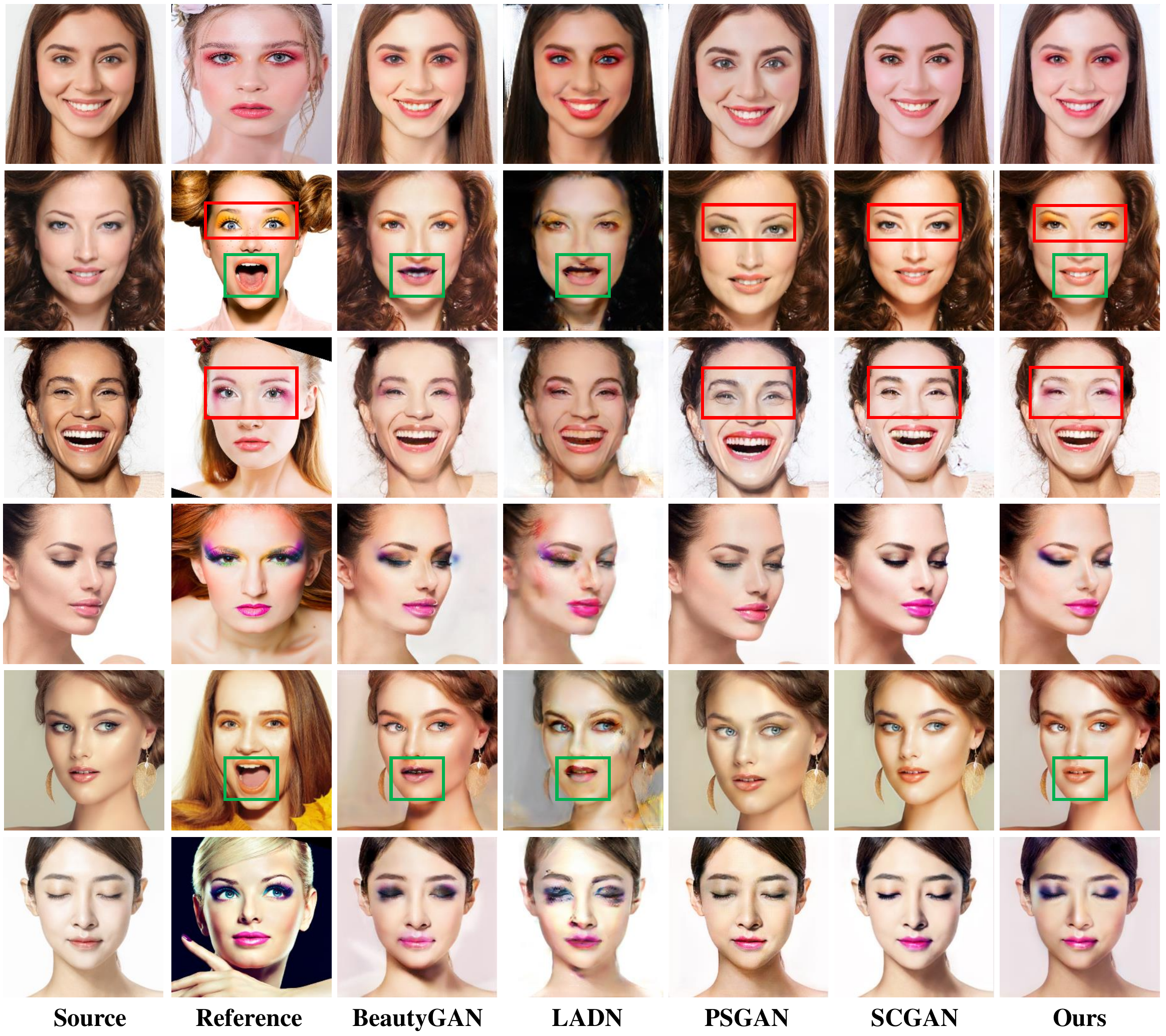}
	\end{center}
	\caption{Qualitative comparisons with previous state-of-the-art makeup transfer methods on the MC dataset. Our method achieves better accuracy of makeup colors and details in the transferred results, even for images under large spatial misalignment.}
	\label{fig:comparsion}
\end{figure*}
\begin{figure*}[ht]
	\centering
		\includegraphics[width=0.98\textwidth]{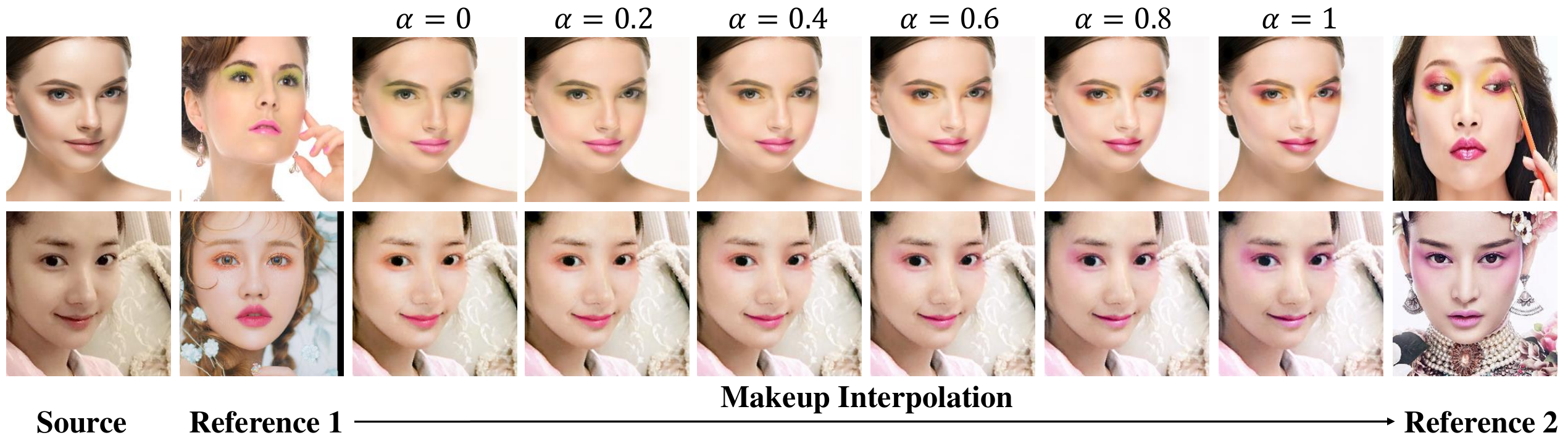}
	\caption{Results of facial makeup interpolation. The makeup styles of the interpolated images continuously transfer from reference 1 to reference 2 by setting the weight $\alpha$ from 0 to 1 in intervals of 0.2.}
	\label{fig:interpolation}
\end{figure*}

  \begin{figure}[tb]
	\centering
		\includegraphics[width=0.85\columnwidth]{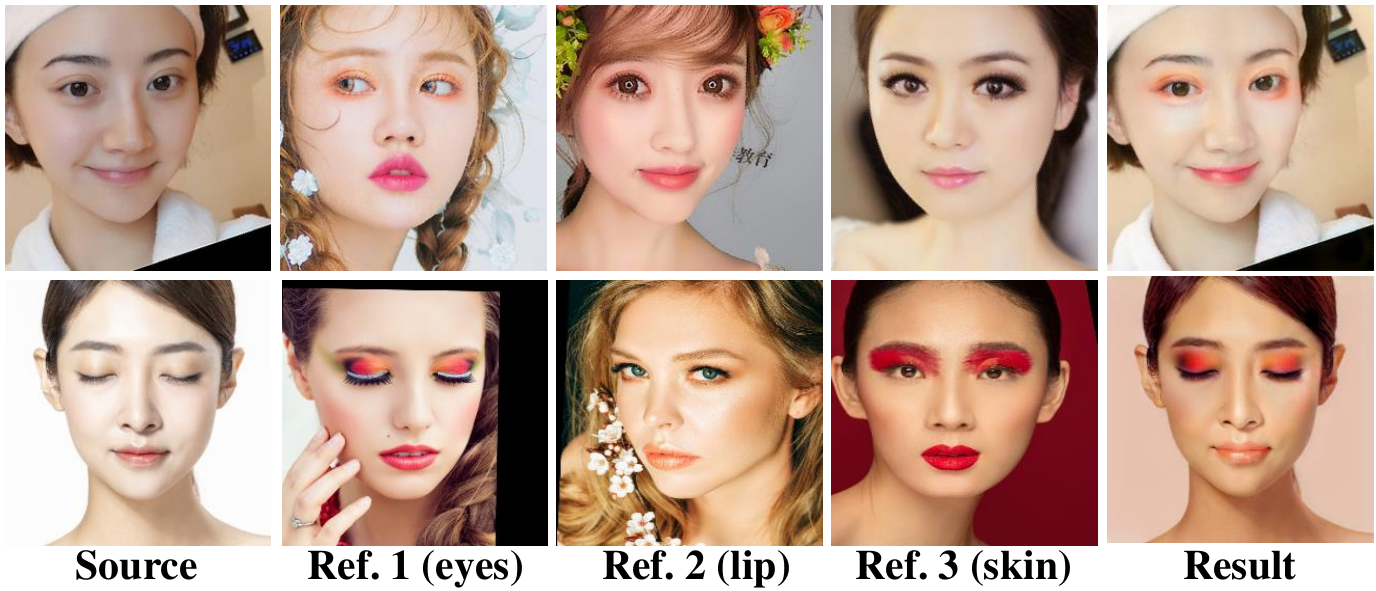}
	\caption{Illustrations of partial makeup transfer.}
	\label{fig:partial}
\end{figure}
\begin{figure}[tb]
	\centering
		\includegraphics[width=0.7\columnwidth]{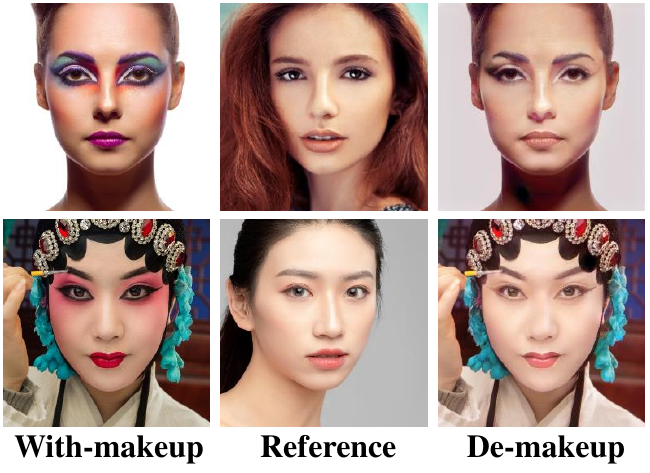}
	\caption{Illustrations of makeup removal results. Our method successfully removes the heavy makeup while preserving the identities of the faces.}
	\label{fig:removal}
\end{figure}

\textbf{Implementation Details.}
For makeup transfer, the models are trained on one 32GB Tesla V100 GPU. 
We set the batch size to 1 and set the training epochs to 50.
For the optimizer, we choose ADAM~\cite{kingma2015adam} with $\beta_{1}=0.5$, $\beta_{2}=0.999$. 
We use a constant learning rate of 0.0002. 
In all experiments, the images are resized to $256 \times 256$. 
We extract features from the $Relu\_4\_1$ layer of pretrained VGG-16 to calculate the perceptual loss. 
The trade-off parameters in \ref{con:totalmloss} are set as $\lambda_{a}^m=1$, $\lambda_{c}^m=10$, $\lambda_{m}^m=1$, $\lambda_{p}^m=10$.
For semantic image synthesis, the models are trained on 4 32GB Tesla V100 GPUs.
We set the batch size to 4 and set the training epochs to 50 for all datasets.
For the optimizer, we choose ADAM~\cite{kingma2015adam} with $\beta_{1}=0$, $\beta_{2}=0.999$. 
Following SPADE~\cite{park2019semantic}, we use constant learning rates of 0.0001 and 0.0004 for the generator and discriminator, respectively. 
We extract features from the output of $Relu\_1\_1$, $Relu\_2\_1$, $Relu\_3\_1$, $Relu\_4\_1$, $Relu\_5\_1$ of pretrained VGG-19 to calculate the perceptual loss.
The trade-off parameters in \ref{con:totalsloss} are set as $\lambda_{a}^s=1$, $\lambda_{fm}^s=10$, $\lambda_{p}^s=10$.

\textbf{Network Architecture Details.}
The number of branches $K$ in the DRAN module is set differently according to the semantic region $R$.
For makeup transfer, 1) For R = skin or lip, K = 2 and $w_{1}=1$, $w_{2}=n/2$,
2) For R = eyes, K = 3 and $w_{1}=1$, $w_{2}=6$, $w_{3}=n/2$,
where n denotes the spatial size of input reference features. 
For semantic image synthesis, 
there are 19 semantic categories in the CelebAMask-HQ dataset and 35 categories in the CityScapes dataset, far exceeding the number in makeup transfer.
Therefore, we only perform multi-branch learning for some most important regions that have more complex details. 
Specifically, the categories with multi-branch on the CelebAMask-HQ dataset is \{skin, hair, earrings\} and on the CityScapes dataset is \{road, building, person, car, motorcycle, bicycle\}. 
As in the makeup transfer task, we set K = 2 and $w_{1}=1$, $w_{2}=n/2$. 
These configurations are empirically shown to be good for both tasks. 
Ablation experiments in Sec.~\ref{ab_k} compare the results of using different numbers of branches.

\textbf{Evaluation Metrics.}
For makeup transfer, we conducted 1) user study to evaluate the subjective perception of makeup transfer effects, 2) PFDM~\cite{organisciak2021makeup} to evaluate the makeup similarity. 
For semantic image synthesis, we assess the generated results from two aspects: 1) segmentation accuracy by mean Intersection-over-Union (mIoU), and pixel accuracy (accu), 2) generation quality by FID~\cite{heusel2017gans}, LPIPS~\cite{zhang2018perceptual}, SSIM~\cite{wang2004image}, RMSE and PSNR.

\subsection{Results on Makeup Transfer}
\subsubsection{Qualitative Comparisons}
We conduct comparisons with four previous state-of-the-art methods, including BeautyGAN~\cite{li2018beautygan}, LADN~\cite{gu2019ladn}, PSGAN~\cite{jiang2020psgan} and SCGAN~\cite{deng2021scgan}. 
For fairness, we use their released codes and pre-trained models.
The results are shown in Fig.~\ref{fig:comparsion}. 
Since BeautyGAN and LADN simply combine the makeup features with the source features by concatenation, they can not handle makeup transfer under spatial misalignment caused by large poses or expressions.
As indicated by green boxes in Fig.~\ref{fig:comparsion}, their transferred results in mouth regions have some artifacts due to the spatial misalignment problem. 
PSGAN employs pixel-to-region attention to align the makeup features with the source features, and then utilizes spatial-adaptive normalization to perform makeup transfer. 
However, the attention map of PSGAN is ambiguous and can only transfer a general makeup style. 
SCGAN, an AdaIN-based method, encodes the makeup information into channel-wise normalization parameters and thus inevitably leads to loss in spatial details. 
\begin{table}[t]
	\centering
		\caption{User preference study on makeup transfer. ``MT Set" and ``MC Set" represent the Makeup Transfer test set and the Makeup-Complex dataset.}
    \resizebox{\columnwidth}{!}{
	\begin{tabular}{cccccc}
		\toprule   
		& BeautyGAN & LADN & PSGAN & SCGAN & Ours \\
		\hline  
		MT Set &$8.63\%$ & $1.00\%$ & $9.25\%$ & $\underline{9.50\%}$ & $\mathbf{71.63\%}$ \\
		MC Set & $\underline{17.88\%} $ & $1.88\%$ & $7.75\% $ & $1.75\%$ & $\mathbf{70.75\%}$\\
		\bottomrule
		\end{tabular}
		}
\label{tab:table1}
\end{table}
\begin{table}[t]
	\centering
	\caption{Comparisons with other methods on makeup similarity using the Proportionate Face Distance Metric (PFDM). The lower the better.}
	\resizebox{\columnwidth}{!}{
	\label{tab:table2}
		\begin{tabular}{cccccc}
			\toprule   
			PFDM$\downarrow$& BeautyGAN & LADN & PSGAN & SCGAN & Ours \\
			\hline   
			MT Set &$0.071$ & $0.102$ & $0.088$ & $\underline{0.064}$ & $\mathbf{0.056}$ \\
			MT-Complex Set &$0.071$ & $0.099$ & $0.088$ & $\underline{0.065}$ & $\mathbf{0.057}$ \\ 
			MT-Simple Set &$0.070$ & $0.102$ & $0.088$ & $\underline{0.064}$ & $\mathbf{0.056}$ \\
			MC Set &$\underline{0.084}$ & $0.115$ & $0.132$ & $0.093$ & $\mathbf{0.069}$ \\
			\bottomrule
			\end{tabular}
			}
\end{table}
\begin{table}[!t]

    \caption{Comparisons of different interpolation methods (e.g., nearest, bilinear, bicubic) on makeup similarity using the Proportionate Face Distance Metric (PFDM). The lower the better.}
    \label{tab:table3}
    \begin{center}
    \resizebox{0.8\columnwidth}{!}{
        \begin{tabular}{cccccc}
            \toprule   
            PFDM$\downarrow$& nearest & bilinear & bicubic \\
            \hline   
            MT Set &$0.056$ & $0.057$ & $0.056$ \\
            MT-Complex Set &$0.057$ & $0.057$ & $0.056$ \\
            MT-Simple Set &$0.056$ & $0.056$ & $0.055$ \\
            MC Set &$0.069$ & $0.069$ & $0.068$ \\
            \bottomrule
            \end{tabular}}
            \end{center}
\end{table}
As indicated by the red boxes in Fig.~\ref{fig:comparsion}, PSGAN and SCGAN only transfer a general style, losing many local details.
By contrast, our approach adaptively fuses coarse-to-fine makeup features in a region-wise manner, and it can achieve accurate transfer of color and makeup details even for images with different poses and expressions.

\begin{figure*}[t]
	\centering
	\includegraphics[width=0.9\textwidth]{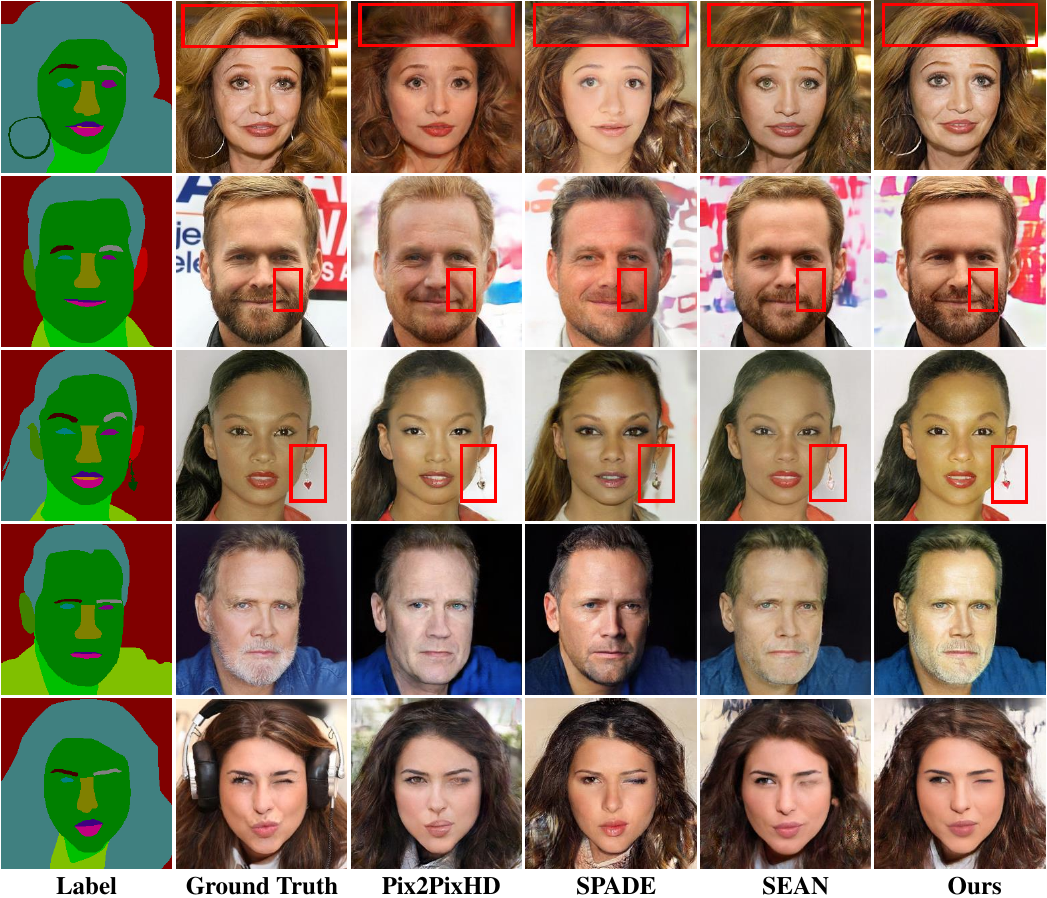}
	\caption{Visualization of semantic image synthesis results on CelebAMask-HQ dataset. Our method generates more accurate details, \emph{e.g.}, facial expressions and hairs.}
	\label{fig:sis_comp_face}
\end{figure*}
	
	\begin{figure*}[t]
		\centering
			\includegraphics[width=0.85\textwidth]{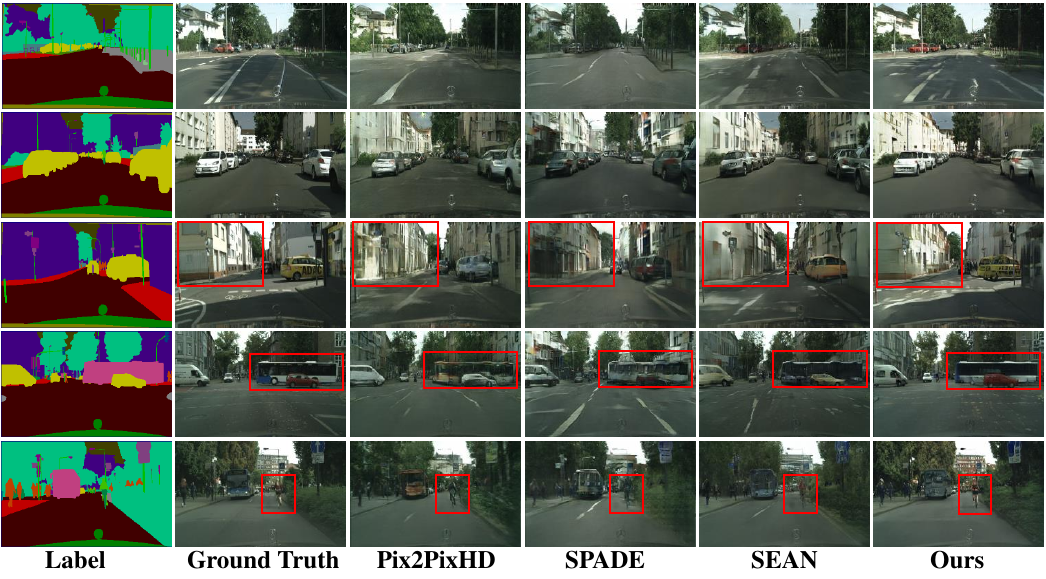}
		\caption{Visualization of semantic image synthesis results on the CityScapes dataset.}
		\label{fig:sis_comp_car}
	\end{figure*}
	
	\begin{figure}[t]
		\centering
			\includegraphics[width=0.85\columnwidth]{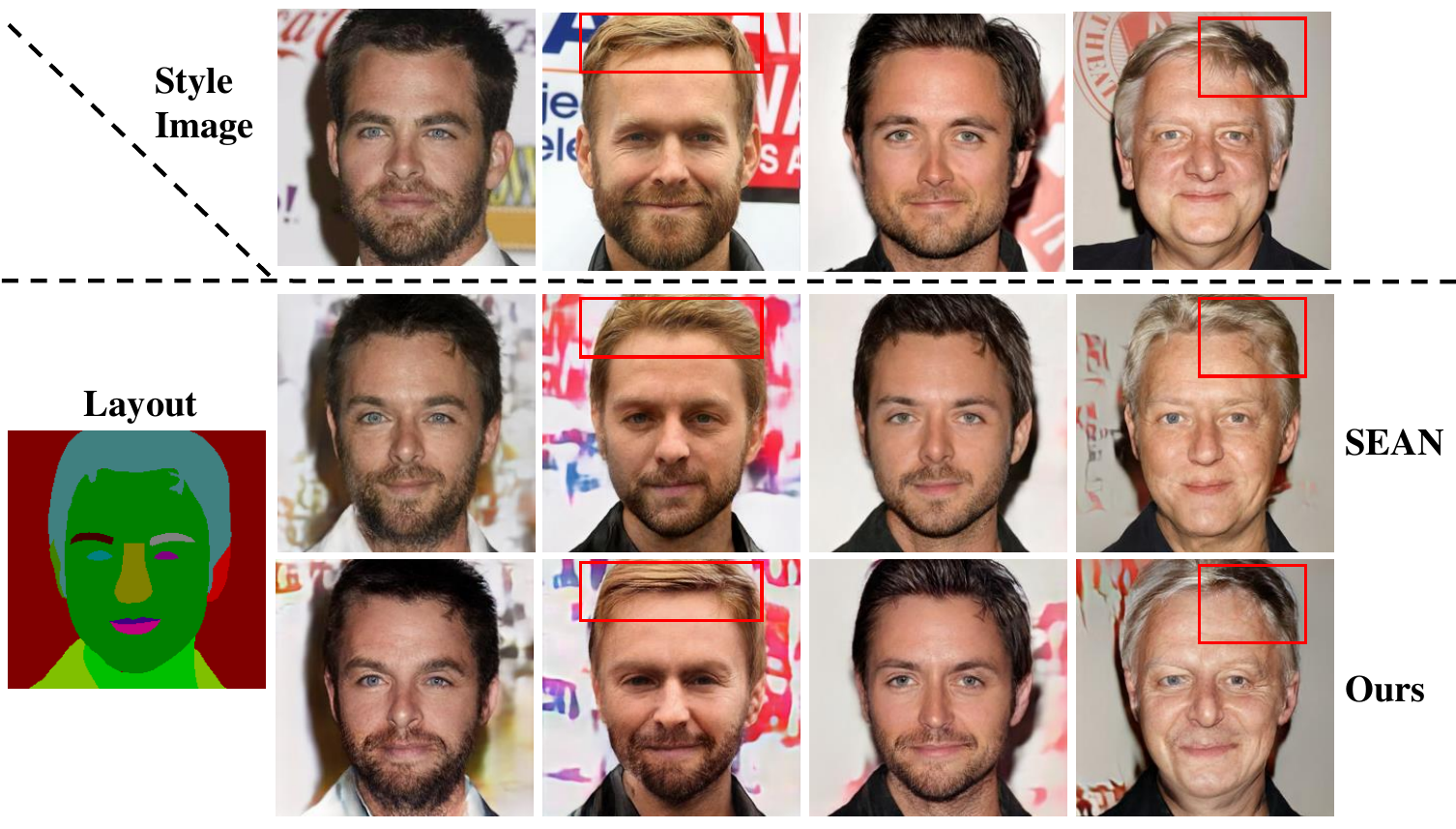}
		\caption{Illustrations of various edits by using different style images. Compared to SEAN, our method preserves more accurate spatial texture information of the style images. For example, the orientation of the hair and the black-and-white colors of the hair in the style images are preserved successfully~(highlighted in red boxes).}
		\label{fig:sty_man}
	\end{figure}
	
	\begin{figure}[t]
		\centering
			\includegraphics[width=0.85\columnwidth]{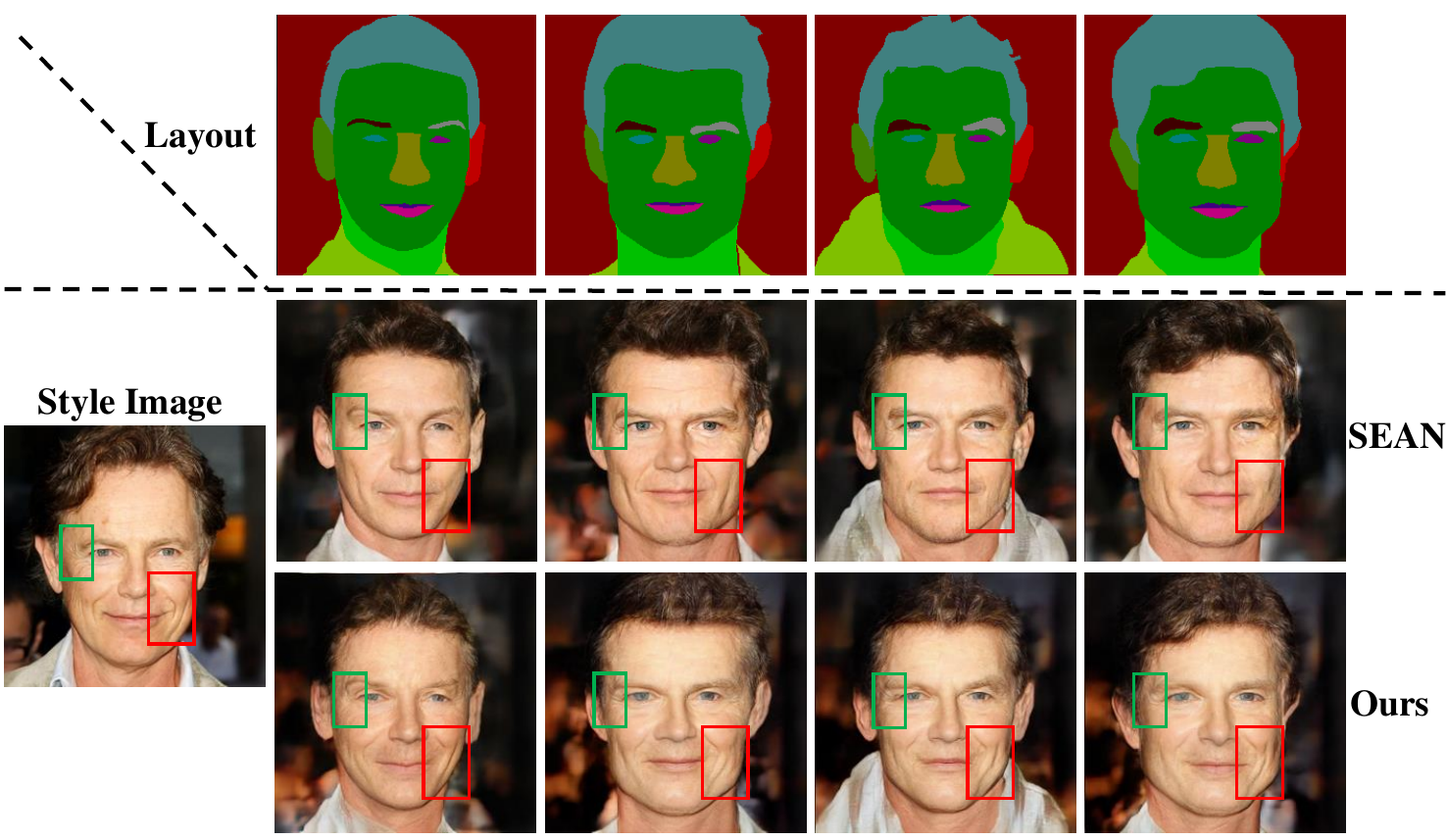}
		\caption{Illustrations of various edits by transferring the same style to different semantic layouts. Compared to SEAN, our generated results have more stylistic similarity with the style image, such as facial expressions and wrinkle details~(highlighted in the red and green boxes).}
		\label{fig:sty_woman}
	\end{figure}
	\begin{figure*}[t]
		\centering
			\includegraphics[width=0.85\textwidth]{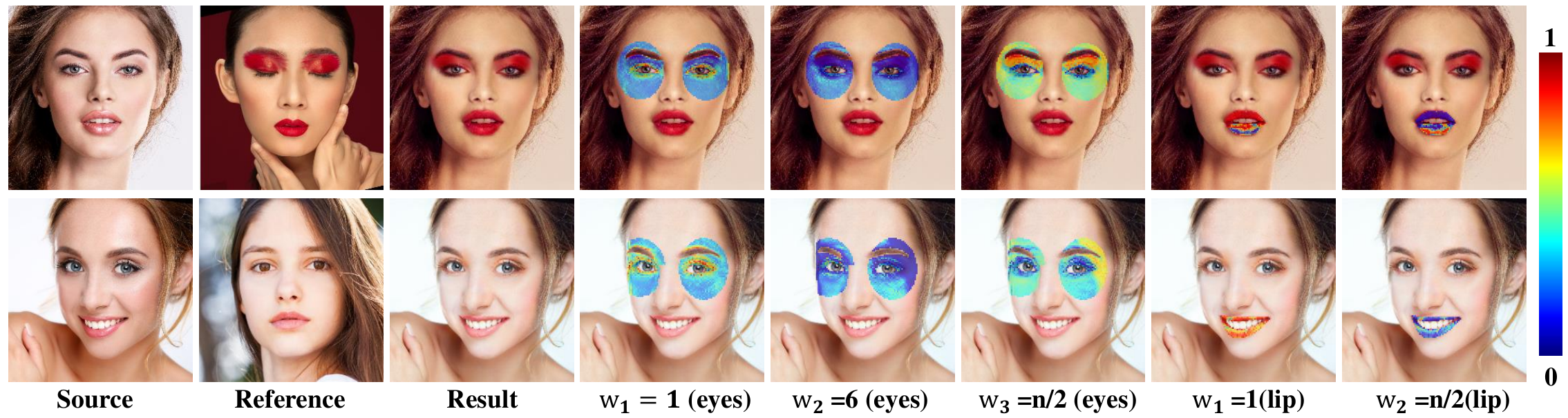}
		\caption{Visualizations of region-wise weighting maps by combing the K branches in DRAN.}
		\label{fig:visulization}
	\end{figure*}
	\begin{table*}[t]
		\centering
		\caption{Quantitative comparison with other methods on the CelebAMask-HQ dataset and the CityScapes dataset. Our method outperforms other methods in FID, LPIPS, SSIM, RMSE, PSNR on all datasets. }
		\resizebox{\textwidth}{!}{
			\begin{tabular}{cccccccc|ccccccc}
			\toprule
			\multicolumn{1}{c}{Method} & \multicolumn{7}{c|}{CelebAMask-HQ}                                                                                  & \multicolumn{7}{c}{CityScapes}                                                                    \\ 
			\cline{2-15} 
			&mIOU$\uparrow$&accu$\uparrow$&FID$\downarrow$&LPIPS$\downarrow$&SSIM$\uparrow$&RMSE$\downarrow$&PSNR$\uparrow$      &mIOU$\uparrow$&accu$\uparrow$&FID$\downarrow$&LPIPS$\downarrow$&SSIM$\uparrow$ &RMSE$\downarrow$&PSNR$\uparrow$\\ 
			\hline
			Pix2PixHD & 73.08&94.59&37.45&0.2856&0.65&0.15&16.54&46.69&90.42&55.49&0.3602&0.64&0.14&17.61\\
			SPADE  & \textbf{76.94} & \textbf{95.28}&22.63&0.3044&0.63&0.18&15.17&\underline{52.14}&\underline{91.95}&60.96&0.4010&0.63&0.17&15.94\\
			SEAN   &\underline{75.94}&\underline{95.03}&\underline{20.45}&\underline{0.2077}&\underline{0.70}&\underline{0.12}&\underline{18.48}&\textbf{54.50}&\textbf{92.37}&\underline{52.88}&\underline{0.3360}&\underline{0.67}&\underline{0.13}&\underline{17.92}\\
			Ours   &74.20& 94.40&\textbf{15.84}&\textbf{0.1490}&\textbf{0.76}&\textbf{0.10}&\textbf{20.53}&51.38&91.65&\textbf{48.92}&\textbf{0.2599}&\textbf{0.73}&\textbf{0.10}&\textbf{19.97}\\
			\bottomrule
			\end{tabular}
			}
		\label{tab:sis_tab}
	\end{table*}
\begin{figure}[t]
	\centering
		\includegraphics[width=0.85\columnwidth]{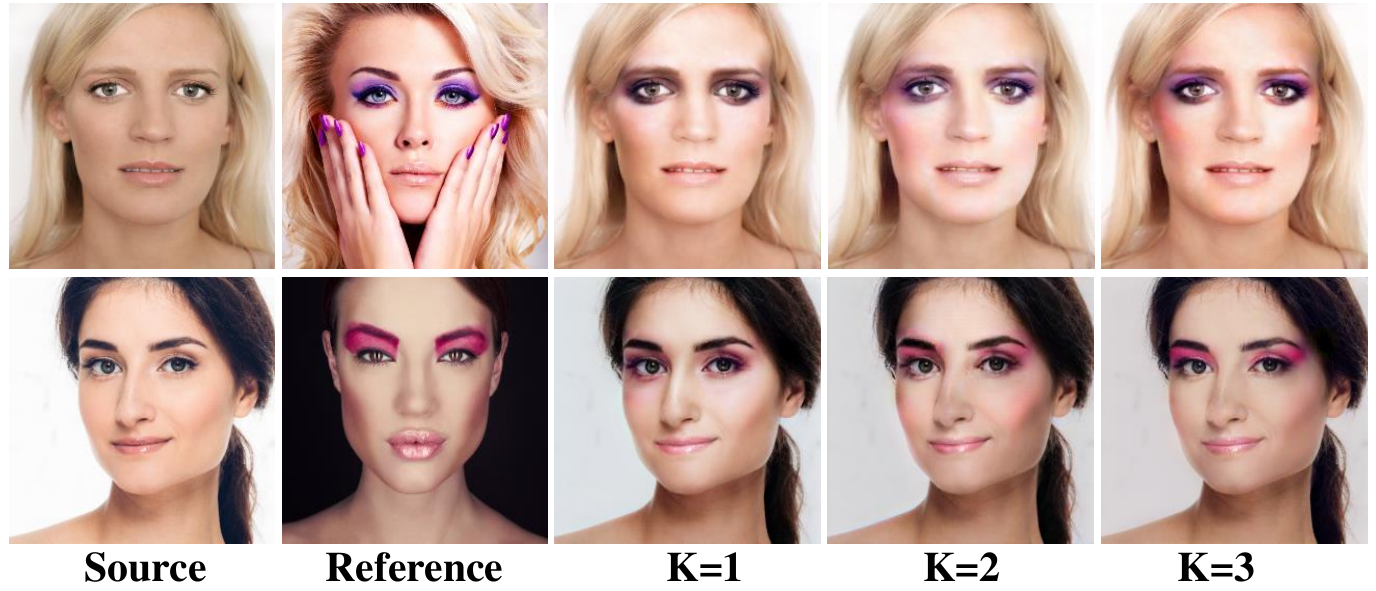}
	\caption{Comparison results using different pyramid levels in SAPP.}
	\label{fig:ablation}
\end{figure}
\begin{figure}[t]
	\centering
		\includegraphics[width=0.85\columnwidth]{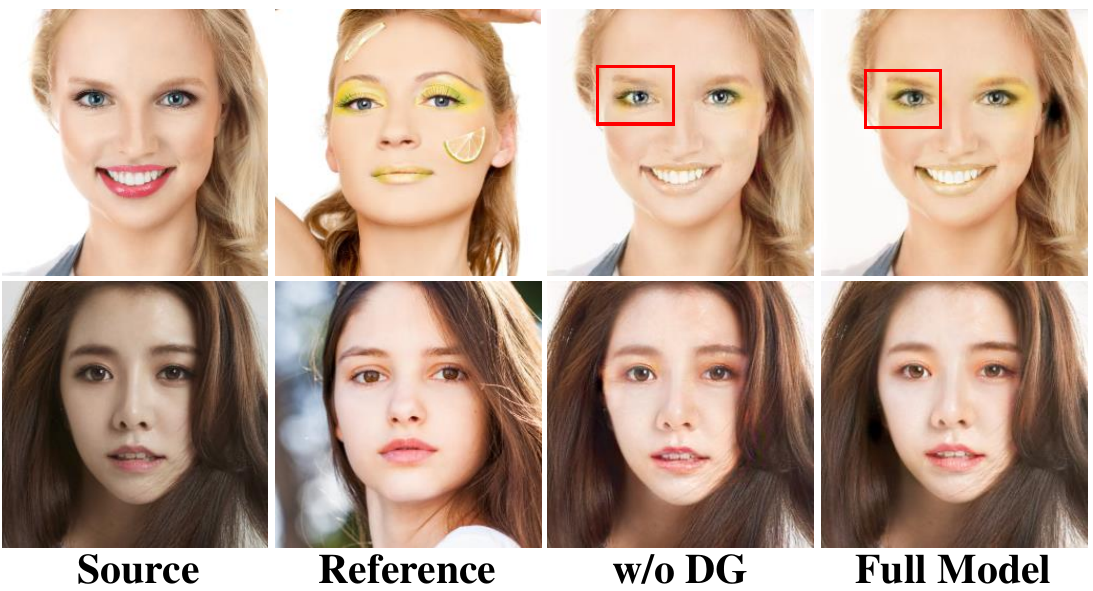}
	\caption{The results of ablation study on Dynamic Gating, where ``w/o DG'' represents removing the Dynamic Gating and directly averaging multiple SAPP branches instead.}
	\label{fig:ablation_dg}
\end{figure}
\subsubsection{Quantitative Comparisons}
We conduct user studies on each dataset and obtain 800 answers for each study. 
The evaluation details are as follows. 
Firstly, we randomly select 20 generated images from the MT and MC test results to form two user studies, respectively. 
There are 40 participants who joined the user studies. 
For each user study, the participants are asked to choose the best from all methods considering the visual quality and the similarity with reference makeup styles. 
As observed in Table.~\ref{tab:table1}, our method outperforms the rest with a large margin on all tests, which demonstrates the effectiveness of DRAN.

To further evaluate the makeup similarity, we use the Proportionate Face Distance Metric (PFDM) proposed in~\cite{organisciak2021makeup}.
PFDM computes the color histogram for each face region and then calculates the L1 distance between the histograms of corresponding regions.
As shown in Table.~\ref{tab:table2}, our method outperforms the other methods on both the MT dataset and the MC dataset. 
We also evaluate the methods on the simple and complex subsets of MT respectively. 
Our method also surpasses other methods in both the simple and complex makeup styles.

Moreover, to verify how different interpolation methods (\emph{e.g.}, nearest, bilinear, bicubic) used for resizing affect the resulting modulation parameters, we conduct quantitative comparison experiments and show the results in Table.~\ref{tab:table3}. The results show that using bicubic interpolation to resize the modulation parameters can bring a slight improvement on PFDM results, since it is more accurate than the other two interpolation methods. However, in general, different interpolation methods have very little impact on the performance of our proposed method, which can still maintain the best performance compared to previous state-of-the-art works.

\subsubsection{Interpolated and Partial Makeup Transfer}
Firstly, facial makeup interpolation is achieved by introducing an interpolation weight $\alpha$ to the feature space:
\begin{equation}
    \begin{aligned}
        V_{inter} = \alpha \times V_{ref1} + (1-\alpha) \times V_{ref2} \text{,}
    \end{aligned}
\end{equation}
where $V_{ref1}$ and $V_{ref2}$ are features extracted from two reference images.
As illustrated in Fig.~\ref{fig:interpolation}, our method achieves smooth and appealing interpolation results between two reference makeups.
The results are also natural even when Reference 2 (shown in the second row) has heavy eye-shadow and cheek color.
Secondly, partial makeup transfer can be easily achieved using multiple reference images with DRAN. 
As shown in Fig.~\ref{fig:partial}, we extract the makeup style of eyes, lips, and skin respectively from Ref. 1, Ref. 2, and Ref. 3 and obtain a new mixed makeup style.
\subsubsection{Makeup Removal}
Our method successfully removes the complex makeup by using non-makeup reference images, 
as shown in Fig.~\ref{fig:removal}. To verify the identity preservation performance when performing makeup removal, we use SFace~\cite{zhong2021sface} to calculate the face verification rates between the face before and after makeup removal on Makeup Transfer dataset and Makeup Complex dataset. The average verification rate of our method on Makeup Transfer dataset is 0.9902, on Makeup Complex dataset is 1.0000, which demonstrate that our method (DRAN) can not only remove different makeup styles (simple or heavy makeup) but also preserve the identities well between the with-makeup faces and the de-makeup faces.

\subsection{Results on Semantic Image Synthesis}
\subsubsection{Qualitative Comparisons}
We compare our method with three previous state-of-the-art methods, Pix2PixHD~\cite{wang2018high}, SPADE~\cite{park2019semantic} and SEAN~\cite{zhu2020sean}.
The results on the CelebAMask-HQ dataset are shown in Fig.~\ref{fig:sis_comp_face}.
Pix2PixHD inputs the reference image through an encoder, and then use average pooling to compute the average features in each region.
The average features are then broadcast to the entire region and decoded to obtain the final reconstruction result.
Therefore, it has difficulty to preserve the original facial identities details in each region, which can be seen in Fig.~\ref{fig:sis_comp_face}. 
SPADE encodes the reference image and the semantic mask in the whole spatial space, instead of by each region, thus it is hard to preserve local details~(\emph{e.g.}, the 4th column in Fig.~\ref{fig:sis_comp_face}). 
SEAN, based on region-wise normalization, can preserve the facial identities of the reference image, but cannot well handle details in each region, such as earrings and facial wrinkles. 
As highlighted by the red boxes in Fig.~\ref{fig:sis_comp_face}, our proposed method yields the most precise results in details, especially the orientation of hairs, facial wrinkles, and the color of earrings. 
The comparison results on the CityScapes dataset are illustrated in Fig.~\ref{fig:sis_comp_car}. 
Due to DRAN's ability in modeling fine-grained styles, our method yields the most precise results for detailed textures, like the texture of buildings, vehicles and persons~(highlighted in the red boxes). 
Moreover, we can achieve various edits by using different style images and different semantic layouts as input. 
The results are shown in Fig.~\ref{fig:sty_man} and Fig.~\ref{fig:sty_woman}.
Compared with the state-of-the-art method SEAN, our method generates more fine-grained details, such as facial textures and local hair details.

\subsubsection{Quantitative Comparisons}
As shown in Table.\ref{tab:sis_tab}, our method achieves lower FID and LPIPS than other methods. 
It indicates our method generates realistic results with fewer artifacts.
Moreover, our method has the best performance in terms of the similarity metrics SSIM, RMSE and PSNR on all datasets, which indicate better reconstruction accuracy.

\subsection{Analysis and Ablation Studies}

\subsubsection{Number of the SAPP branches}\label{ab_k}
We conduct experimental analysis on SAPP branches with different pyramid levels.
Specifically, we use three settings, 1) when K = 1, $w_{1}=1$; 2) when K = 2, $w_{1}=1$, $w_{2}=n/2$; 3) when K = 3, $w_{1}=1$, $w_{2}=6$, $w_{3}=n/2$. The qualitative results are shown in Fig.~\ref{fig:ablation}.
When the number of branches is set to 1, our method only transfers the general style while losing fine-grained details.
When increasing the number of branches and introducing finer pyramid levels, performance improvement can be observed in transferring fine-grained details. 
It verifies that multi-level SAPP branches capture multiple granularities of styles, which is the key to improving the results.

\subsubsection{Effectiveness of the Dynamic Gating}
First, we verify the effectiveness of the Dynamic Gating by some visualization experiments on makeup transfer. 
The weighting maps of each branch are visualized in Fig.~\ref{fig:visulization}, which are obtained by showing the $\alpha_R^k$ in Equation.~\ref{equ:dg}.
For the eyes region, we notice that the reference image with more details~(the first row) is given more weights on the fine branch ($w_{3}=n/2$) than on the coarse branch ($w_{1}=1$). 
On the contrary, for the reference image with a general makeup style~(the second row), the dynamic gate reduces the weights on the fine branch ($w_{3}=n/2$) and increases the weights on the coarse branch ($w_{1}=1$). 
For the lip region, the two reference images both have a general makeup style, and the coarse branch ($w_{1}=1$) has higher weights than the fine branch ($w_{2}=n/2$).
It demonstrates that the dynamic gate in DRAN can adaptively adjust the weights of different branches according to the different levels of details in each specific region, which is consistent with our motivation of design.

Second, we conduct ablation study by replacing the Dynamic Gating with direct averaging of multiple SAPP branches. 
The qualitative results are shown in Fig.~\ref{fig:ablation_dg}.
Under the configuration of ``w/o DG'', \emph{i.e.}, without Dynamic Gating, the method still has the ability to transfer general makeup styles~(the second row), but cannot transfer well on reference images with complex details~(the first row), which is consistent with the visualization results above.
For reference images with rich details, it is more important to learn coarse-to-fine style representation adaptively, which further demonstrates the importance of Dynamic Gating to adaptively ensemble multiple style levels.

\section{Conclusion}
In this paper, we proposed a simple yet effective method, Detailed Region-Adaptive Normalization, for conditional image synthesis. 
It is composed of Spatiality-aware Pyramid Pooling and Dynamic Gating. 
Both work together to effectively encode styles from conditional inputs, and learn coarse-to-fine visual style representation adaptively.
Since DRAN is a general module, we apply DRAN to two typical tasks, makeup transfer and semantic image synthesis,
and observe remarkable performance gains in qualitative and quantitative experiments.
In future work, we will explore its usage and possible variants in other image generation tasks, such as style transfer, and text-to-image generation.

\section{Acknowledgements} 
This work is supported by the National Key Research and Development Program of China under Grant No. 2021YFC3320103, the National Natural Science Foundation of China (NSFC) under Grants 62272460, U19B2038, Beijing Natural Science Foundation under Grant No. 4232037, a grant from Young Elite Scientists Sponsorship Program by CAST (YESS), and CAAI-Huawei MindSpore Open Fund.

\ifCLASSOPTIONcaptionsoff
  \newpage
\fi

\bibliographystyle{IEEEtran}
\bibliography{main}

\begin{thebibliography}{10}
\providecommand{\url}[1]{#1}
\csname url@samestyle\endcsname
\providecommand{\newblock}{\relax}
\providecommand{\bibinfo}[2]{#2}
\providecommand{\BIBentrySTDinterwordspacing}{\spaceskip=0pt\relax}
\providecommand{\BIBentryALTinterwordstretchfactor}{4}
\providecommand{\BIBentryALTinterwordspacing}{\spaceskip=\fontdimen2\font plus
\BIBentryALTinterwordstretchfactor\fontdimen3\font minus
  \fontdimen4\font\relax}
\providecommand{\BIBforeignlanguage}[2]{{%
\expandafter\ifx\csname l@#1\endcsname\relax
\typeout{** WARNING: IEEEtran.bst: No hyphenation pattern has been}%
\typeout{** loaded for the language `#1'. Using the pattern for}%
\typeout{** the default language instead.}%
\else
\language=\csname l@#1\endcsname
\fi
#2}}
\providecommand{\BIBdecl}{\relax}
\BIBdecl

\bibitem{goodfellow2014generative}
I.~Goodfellow, J.~Pouget-Abadie, M.~Mirza, B.~Xu, D.~Warde-Farley, S.~Ozair,
  A.~Courville, and Y.~Bengio, ``Generative adversarial nets,'' in
  \emph{Advances in Neural Information Processing Systems (NeurIPS)}, vol.~27,
  2014.

\bibitem{karras2017progressive}
T.~Karras, T.~Aila, S.~Laine, and J.~Lehtinen, ``Progressive growing of gans
  for improved quality, stability, and variation,'' in \emph{International
  Conference on Learning Representations (ICLR)}, 2018.

\bibitem{brock2018large}
A.~Brock, J.~Donahue, and K.~Simonyan, ``Large scale gan training for high
  fidelity natural image synthesis,'' in \emph{International Conference on
  Learning Representations (ICLR)}, 2019.

\bibitem{karras2019style}
T.~Karras, S.~Laine, and T.~Aila, ``A style-based generator architecture for
  generative adversarial networks,'' in \emph{Proceedings of the IEEE
  Conference on Computer Vision and Pattern Recognition (CVPR)}, 2019, pp.
  4401--4410.

\bibitem{choi2018stargan}
Y.~Choi, M.~Choi, M.~Kim, J.-W. Ha, S.~Kim, and J.~Choo, ``Stargan: Unified
  generative adversarial networks for multi-domain image-to-image
  translation,'' in \emph{Proceedings of the IEEE Conference on Computer Vision
  and Pattern Recognition (CVPR)}, 2018, pp. 8789--8797.

\bibitem{he2019attgan}
Z.~He, W.~Zuo, M.~Kan, S.~Shan, and X.~Chen, ``Attgan: Facial attribute editing
  by only changing what you want,'' \emph{IEEE Transactions on Image Processing
  (TIP)}, vol.~28, no.~11, pp. 5464--5478, 2019.

\bibitem{chen2021semantic}
T.~Chen, S.~Wu, X.~Yang, Y.~Xu, and H.-S. Wong, ``Semantic regularized
  class-conditional gans for semi-supervised fine-grained image synthesis,''
  \emph{IEEE Transactions on Multimedia (TMM)}, 2021.

\bibitem{liu2022isf}
Y.~Liu, Y.~Chen, L.~Bao, N.~Sebe, B.~Lepri, and M.~De~Nadai, ``Isf-gan: An
  implicit style function for high-resolution image-to-image translation,''
  \emph{IEEE Transactions on Multimedia (TMM)}, 2022.

\bibitem{zhang2017stackgan}
H.~Zhang, T.~Xu, H.~Li, S.~Zhang, X.~Wang, X.~Huang, and D.~N. Metaxas,
  ``Stackgan: Text to photo-realistic image synthesis with stacked generative
  adversarial networks,'' in \emph{Proceedings of the IEEE International
  Conference on Computer Vision (ICCV)}, 2017, pp. 5907--5915.

\bibitem{hong2018inferring}
S.~Hong, D.~Yang, J.~Choi, and H.~Lee, ``Inferring semantic layout for
  hierarchical text-to-image synthesis,'' in \emph{Proceedings of the IEEE
  Conference on Computer Vision and Pattern Recognition (CVPR)}, 2018, pp.
  7986--7994.

\bibitem{yuan2019ckd}
M.~Yuan and Y.~Peng, ``Ckd: Cross-task knowledge distillation for text-to-image
  synthesis,'' \emph{IEEE Transactions on Multimedia (TMM)}, vol.~22, no.~8,
  pp. 1955--1968, 2019.

\bibitem{li2020exploring}
R.~Li, N.~Wang, F.~Feng, G.~Zhang, and X.~Wang, ``Exploring global and local
  linguistic representations for text-to-image synthesis,'' \emph{IEEE
  Transactions on Multimedia (TMM)}, vol.~22, no.~12, pp. 3075--3087, 2020.

\bibitem{wang2018high}
T.-C. Wang, M.-Y. Liu, J.-Y. Zhu, A.~Tao, J.~Kautz, and B.~Catanzaro,
  ``High-resolution image synthesis and semantic manipulation with conditional
  gans,'' in \emph{Proceedings of the IEEE Conference on Computer Vision and
  Pattern Recognition (CVPR)}, 2018, pp. 8798--8807.

\bibitem{park2019semantic}
T.~Park, M.-Y. Liu, T.-C. Wang, and J.-Y. Zhu, ``Semantic image synthesis with
  spatially-adaptive normalization,'' in \emph{Proceedings of the IEEE
  Conference on Computer Vision and Pattern Recognition (CVPR)}, 2019, pp.
  2337--2346.

\bibitem{huang2020semantic}
J.~Huang, J.~Liao, and S.~Kwong, ``Semantic example guided image-to-image
  translation,'' \emph{IEEE Transactions on Multimedia (TMM)}, vol.~23, pp.
  1654--1665, 2020.

\bibitem{deng2022semantic}
Q.~Deng, Q.~Li, J.~Cao, Y.~Liu, and Z.~Sun, ``Semantic-aware noise driven
  portrait synthesis and manipulation,'' \emph{IEEE Transactions on Multimedia
  (TMM)}, 2022.

\bibitem{li2018beautygan}
T.~Li, R.~Qian, C.~Dong, S.~Liu, Q.~Yan, W.~Zhu, and L.~Lin, ``Beautygan:
  Instance-level facial makeup transfer with deep generative adversarial
  network,'' in \emph{ACM International Conference on Multimedia (ACM MM)},
  2018, pp. 645--653.

\bibitem{chen2019beautyglow}
H.-J. Chen, K.-M. Hui, S.-Y. Wang, L.-W. Tsao, H.-H. Shuai, and W.-H. Cheng,
  ``Beautyglow: On-demand makeup transfer framework with reversible generative
  network,'' in \emph{Proceedings of the IEEE Conference on Computer Vision and
  Pattern Recognition (CVPR)}, 2019, pp. 10\,042--10\,050.

\bibitem{gu2019ladn}
Q.~Gu, G.~Wang, M.~T. Chiu, Y.-W. Tai, and C.-K. Tang, ``Ladn: Local
  adversarial disentangling network for facial makeup and de-makeup,'' in
  \emph{Proceedings of the IEEE International Conference on Computer Vision
  (ICCV)}, 2019, pp. 10\,481--10\,490.

\bibitem{lyu2021sogan}
Y.~Lyu, J.~Dong, B.~Peng, W.~Wang, and T.~Tan, ``Sogan: {3D}-aware shadow and
  occlusion robust gan for makeup transfer,'' in \emph{ACM International
  Conference on Multimedia (ACM MM)}, 2021, pp. 3601--3609.

\bibitem{deng2021scgan}
H.~Deng, C.~Han, H.~Cai, G.~Han, and S.~He, ``Spatially-invariant style-codes
  controlled makeup transfer,'' in \emph{Proceedings of the IEEE Conference on
  Computer Vision and Pattern Recognition (CVPR)}, 2021, pp. 6549--6557.

\bibitem{zhu2020sean}
P.~Zhu, R.~Abdal, Y.~Qin, and P.~Wonka, ``Sean: Image synthesis with semantic
  region-adaptive normalization,'' in \emph{Proceedings of the IEEE Conference
  on Computer Vision and Pattern Recognition (CVPR)}, 2020, pp. 5104--5113.

\bibitem{zhu2020semantically}
Z.~Zhu, Z.~Xu, A.~You, and X.~Bai, ``Semantically multi-modal image
  synthesis,'' in \emph{Proceedings of the IEEE Conference on Computer Vision
  and Pattern Recognition (CVPR)}, 2020, pp. 5467--5476.

\bibitem{tan2021diverse}
Z.~Tan, M.~Chai, D.~Chen, J.~Liao, Q.~Chu, B.~Liu, G.~Hua, and N.~Yu, ``Diverse
  semantic image synthesis via probability distribution modeling,'' in
  \emph{Proceedings of the IEEE Conference on Computer Vision and Pattern
  Recognition (CVPR)}, 2021, pp. 7962--7971.

\bibitem{wang2021image}
Y.~Wang, L.~Qi, Y.-C. Chen, X.~Zhang, and J.~Jia, ``Image synthesis via
  semantic composition,'' in \emph{Proceedings of the IEEE International
  Conference on Computer Vision (ICCV)}, 2021, pp. 13\,749--13\,758.

\bibitem{huang2017arbitrary}
X.~Huang and S.~Belongie, ``Arbitrary style transfer in real-time with adaptive
  instance normalization,'' in \emph{Proceedings of the IEEE International
  Conference on Computer Vision (ICCV)}, 2017, pp. 1501--1510.

\bibitem{huang2018multimodal}
X.~Huang, M.-Y. Liu, S.~Belongie, and J.~Kautz, ``Multimodal unsupervised
  image-to-image translation,'' in \emph{Proceedings of the European Conference
  on Computer Vision (ECCV)}, 2018, pp. 172--189.

\bibitem{wang2018recovering}
X.~Wang, K.~Yu, C.~Dong, and C.~C. Loy, ``Recovering realistic texture in image
  super-resolution by deep spatial feature transform,'' in \emph{Proceedings of
  the IEEE Conference on Computer Vision and Pattern Recognition (CVPR)}, 2018,
  pp. 606--615.

\bibitem{zhang2019self}
H.~Zhang, I.~Goodfellow, D.~Metaxas, and A.~Odena, ``Self-attention generative
  adversarial networks,'' in \emph{International Conference on Machine Learning
  (ICML)}, 2019, pp. 7354--7363.

\bibitem{gatys2016image}
L.~A. Gatys, A.~S. Ecker, and M.~Bethge, ``Image style transfer using
  convolutional neural networks,'' in \emph{Proceedings of the IEEE Conference
  on Computer Vision and Pattern Recognition (CVPR)}, 2016, pp. 2414--2423.

\bibitem{li2016combining}
C.~Li and M.~Wand, ``Combining markov random fields and convolutional neural
  networks for image synthesis,'' in \emph{Proceedings of the IEEE Conference
  on Computer Vision and Pattern Recognition (CVPR)}, 2016, pp. 2479--2486.

\bibitem{li2017demystifying}
Y.~Li, N.~Wang, J.~Liu, and X.~Hou, ``Demystifying neural style transfer,'' in
  \emph{International Joint Conferences on Artificial Intelligence (IJCAI)},
  2017, pp. 2230--2236.

\bibitem{CelebAMask-HQ}
C.-H. Lee, Z.~Liu, L.~Wu, and P.~Luo, ``Maskgan: Towards diverse and
  interactive facial image manipulation,'' in \emph{Proceedings of the IEEE
  Conference on Computer Vision and Pattern Recognition (CVPR)}, 2020, pp.
  5549--5558.

\bibitem{men2020controllable}
Y.~Men, Y.~Mao, Y.~Jiang, W.-Y. Ma, and Z.~Lian, ``Controllable person image
  synthesis with attribute-decomposed gan,'' in \emph{Proceedings of the IEEE
  Conference on Computer Vision and Pattern Recognition (CVPR)}, 2020, pp.
  5084--5093.

\bibitem{ling2021region}
J.~Ling, H.~Xue, L.~Song, R.~Xie, and X.~Gu, ``Region-aware adaptive instance
  normalization for image harmonization,'' in \emph{Proceedings of the IEEE
  Conference on Computer Vision and Pattern Recognition (CVPR)}, 2021, pp.
  9361--9370.

\bibitem{lv2021learning}
Z.~Lv, X.~Li, X.~Li, F.~Li, T.~Lin, D.~He, and W.~Zuo, ``Learning semantic
  person image generation by region-adaptive normalization,'' in
  \emph{Proceedings of the IEEE Conference on Computer Vision and Pattern
  Recognition (CVPR)}, 2021, pp. 10\,806--10\,815.

\bibitem{yu2020region}
T.~Yu, Z.~Guo, X.~Jin, S.~Wu, Z.~Chen, W.~Li, Z.~Zhang, and S.~Liu, ``Region
  normalization for image inpainting,'' in \emph{Thirty-fourth AAAI Conference
  on Artificial Intelligence (AAAI)}, vol.~34, no.~07, 2020, pp.
  12\,733--12\,740.

\bibitem{zhu2017unpaired}
J.-Y. Zhu, T.~Park, P.~Isola, and A.~A. Efros, ``Unpaired image-to-image
  translation using cycle-consistent adversarial networks,'' in
  \emph{Proceedings of the IEEE International Conference on Computer Vision
  (ICCV)}, 2017, pp. 2223--2232.

\bibitem{chang2018pairedcyclegan}
H.~Chang, J.~Lu, F.~Yu, and A.~Finkelstein, ``Pairedcyclegan: Asymmetric style
  transfer for applying and removing makeup,'' in \emph{Proceedings of the IEEE
  Conference on Computer Vision and Pattern Recognition (CVPR)}, 2018, pp.
  40--48.

\bibitem{kingma2018glow}
D.~P. Kingma and P.~Dhariwal, ``Glow: Generative flow with invertible 1x1
  convolutions,'' in \emph{Advances in Neural Information Processing Systems
  (NeurIPS)}, vol.~31, 2018.

\bibitem{jiang2020psgan}
W.~Jiang, S.~Liu, C.~Gao, J.~Cao, R.~He, J.~Feng, and S.~Yan, ``Psgan: Pose and
  expression robust spatial-aware gan for customizable makeup transfer,'' in
  \emph{Proceedings of the IEEE Conference on Computer Vision and Pattern
  Recognition (CVPR)}, 2020, pp. 5194--5202.

\bibitem{isola2017image}
P.~Isola, J.-Y. Zhu, T.~Zhou, and A.~A. Efros, ``Image-to-image translation
  with conditional adversarial networks,'' in \emph{Proceedings of the IEEE
  Conference on Computer Vision and Pattern Recognition (CVPR)}, 2017, pp.
  1125--1134.

\bibitem{ioffe2015batch}
S.~Ioffe and C.~Szegedy, ``Batch normalization: Accelerating deep network
  training by reducing internal covariate shift,'' in \emph{International
  Conference on Machine Learning (ICML)}, 2015, pp. 448--456.

\bibitem{ulyanov2016instance}
D.~Ulyanov, A.~Vedaldi, and V.~Lempitsky, ``Instance normalization: The missing
  ingredient for fast stylization,'' \emph{arXiv preprint arXiv:1607.08022},
  2016.

\bibitem{ba2016layer}
J.~L. Ba, J.~R. Kiros, and G.~E. Hinton, ``Layer normalization,'' \emph{arXiv
  preprint arXiv:1607.06450}, 2016.

\bibitem{wu2018group}
Y.~Wu and K.~He, ``Group normalization,'' in \emph{Proceedings of the European
  Conference on Computer Vision (ECCV)}, 2018, pp. 3--19.

\bibitem{tan2021efficient}
Z.~Tan, D.~Chen, Q.~Chu, M.~Chai, J.~Liao, M.~He, L.~Yuan, G.~Hua, and N.~Yu,
  ``Efficient semantic image synthesis via class-adaptive normalization,''
  \emph{IEEE Transactions on Pattern Analysis and Machine Intelligence
  (TPAMI)}, 2021.

\bibitem{krizhevsky2012imagenet}
A.~Krizhevsky, I.~Sutskever, and G.~E. Hinton, ``Imagenet classification with
  deep convolutional neural networks,'' in \emph{Advances in Neural Information
  Processing Systems (NeurIPS)}, vol.~25, 2012.

\bibitem{he2015spatial}
K.~He, X.~Zhang, S.~Ren, and J.~Sun, ``Spatial pyramid pooling in deep
  convolutional networks for visual recognition,'' \emph{IEEE Transactions on
  Pattern Analysis and Machine Intelligence (TPAMI)}, vol.~37, no.~9, pp.
  1904--1916, 2015.

\bibitem{johnson2016perceptual}
J.~Johnson, A.~Alahi, and L.~Fei-Fei, ``Perceptual losses for real-time style
  transfer and super-resolution,'' in \emph{Proceedings of the European
  Conference on Computer Vision (ECCV)}, 2016, pp. 694--711.

\bibitem{liu2015deep}
Z.~Liu, P.~Luo, X.~Wang, and X.~Tang, ``Deep learning face attributes in the
  wild,'' in \emph{Proceedings of the IEEE International Conference on Computer
  Vision (ICCV)}, 2015, pp. 3730--3738.

\bibitem{cordts2016cityscapes}
M.~Cordts, M.~Omran, S.~Ramos, T.~Rehfeld, M.~Enzweiler, R.~Benenson,
  U.~Franke, S.~Roth, and B.~Schiele, ``The cityscapes dataset for semantic
  urban scene understanding,'' in \emph{Proceedings of the IEEE Conference on
  Computer Vision and Pattern Recognition (CVPR)}, 2016, pp. 3213--3223.

\bibitem{kingma2015adam}
D.~P. Kingma and J.~Ba, ``Adam: A method for stochastic optimization,'' in
  \emph{International Conference on Learning Representations (ICLR)}, 2015.

\bibitem{organisciak2021makeup}
D.~Organisciak, E.~S. Ho, and H.~P. Shum, ``Makeup style transfer on
  low-quality images with weighted multi-scale attention,'' in
  \emph{International Conference on Pattern Recognition (ICPR)}, 2021.

\bibitem{heusel2017gans}
M.~Heusel, H.~Ramsauer, T.~Unterthiner, B.~Nessler, and S.~Hochreiter, ``Gans
  trained by a two time-scale update rule converge to a local nash
  equilibrium,'' in \emph{Advances in Neural Information Processing Systems
  (NeurIPS)}, vol.~30, 2017.

\bibitem{zhang2018perceptual}
R.~Zhang, P.~Isola, A.~A. Efros, E.~Shechtman, and O.~Wang, ``The unreasonable
  effectiveness of deep features as a perceptual metric,'' in \emph{Proceedings
  of the IEEE Conference on Computer Vision and Pattern Recognition (CVPR)},
  2018, pp. 586--595.

\bibitem{wang2004image}
Z.~Wang, A.~C. Bovik, H.~R. Sheikh, and E.~P. Simoncelli, ``Image quality
  assessment: from error visibility to structural similarity,'' \emph{IEEE
  Transactions on Image Processing (TIP)}, vol.~13, no.~4, pp. 600--612, 2004.

\bibitem{zhong2021sface}
Y.~Zhong, W.~Deng, J.~Hu, D.~Zhao, X.~Li, and D.~Wen, ``Sface:
  Sigmoid-constrained hypersphere loss for robust face recognition,''
  \emph{IEEE Transactions on Image Processing (TIP)}, vol.~30, pp. 2587--2598,
  2021.

\end{thebibliography}
%
\begin{IEEEbiography}[{\includegraphics[width=1in,height=1.25in,clip,keepaspectratio]{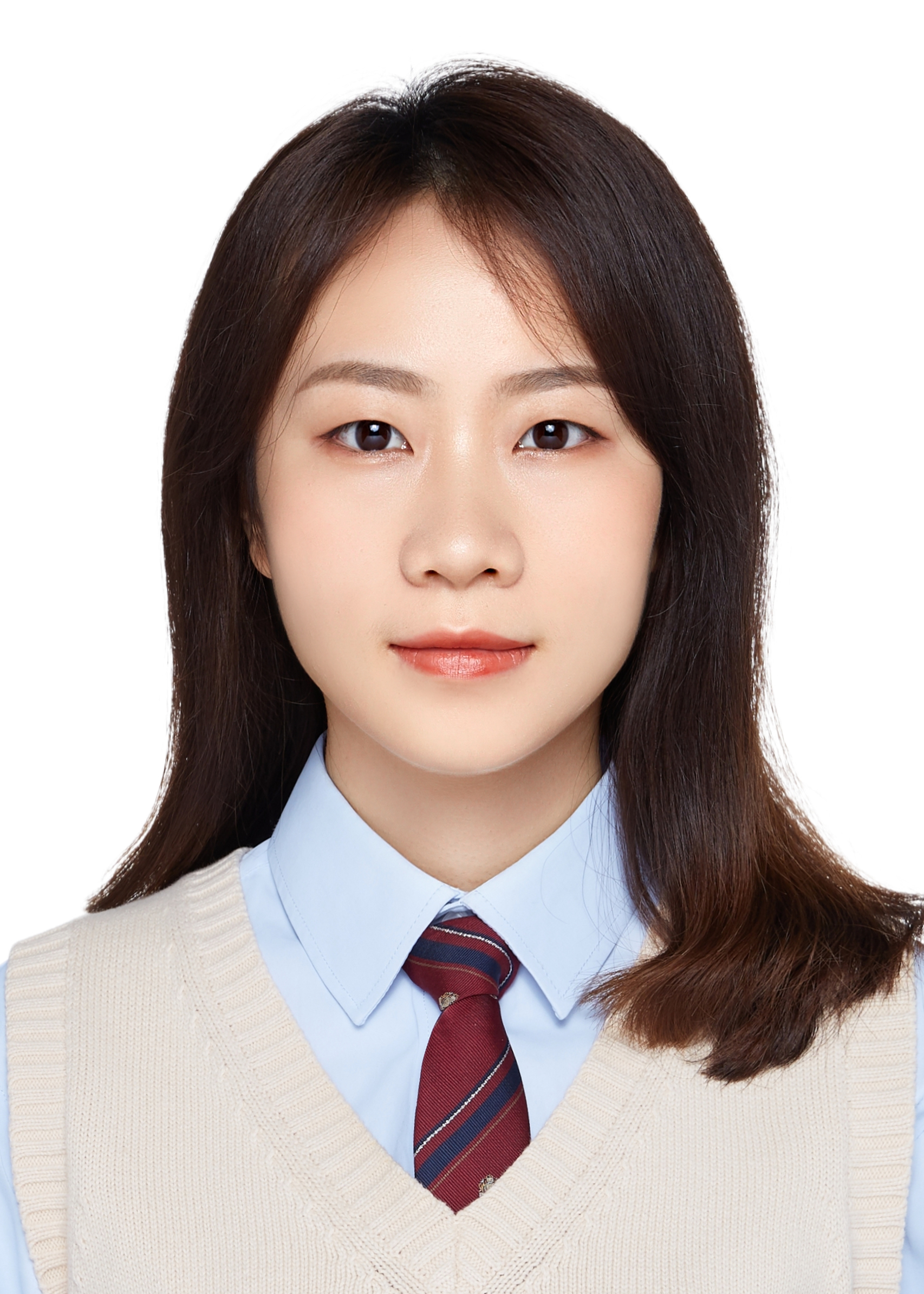}}]{Yueming Lyu}{\space}received B.Eng. degree in Nanjing University of Aeronautics and Astronautics, China in 2019. She is a Ph.D. degree candidate in the Center for Research on Intelligent Perception and Computing (CRIPAC) at the State Key Laboratory of Multimodal Artificial Intelligence Systems, Institute of Automation, Chinese Academy of Sciences, China. Her current research include computer vision, image generation and adversarial learning.

\end{IEEEbiography}

\begin{IEEEbiography}[{\includegraphics[width=1in,height=1.25in,clip,keepaspectratio]{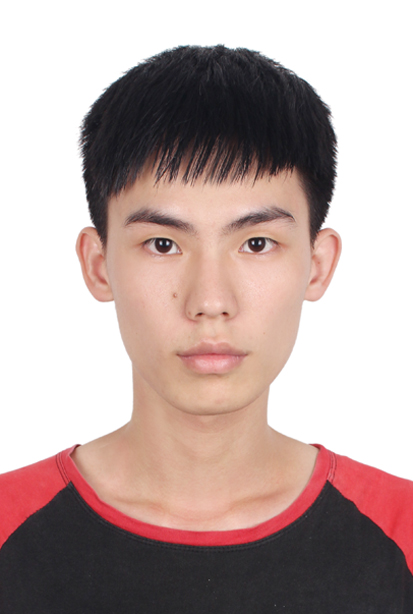}}]{Peibin Chen}{\space}received B.Eng.degree in South China University of Technology and Master degree in Peking University in 2018 and 2021 respectively. Since july 2021, he joined Bytedance. His current research focuses on deep learning, face recognition and semi-supervised learning.

\end{IEEEbiography}

\begin{IEEEbiography}[{\includegraphics[width=1in,height=1.25in,clip,keepaspectratio]{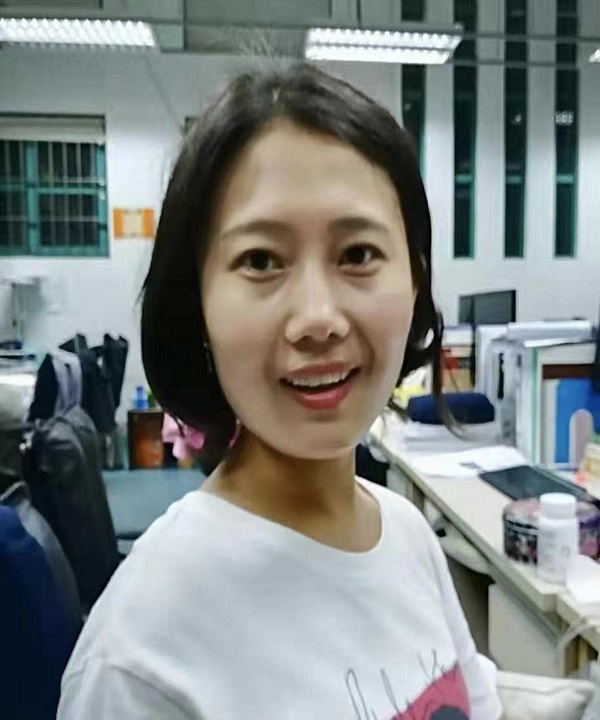}}]{Jingna Sun}{\space}received B.Eng.degree in Jilin University and Master degree in Tsinghua University in 2017 and 2020 respectively. Since july 2020, she joined Bytedance. Her research interests include deep learning, face recognition and image generation.

\end{IEEEbiography}

\begin{IEEEbiography}[{\includegraphics[width=1in,height=1.25in,clip,keepaspectratio]{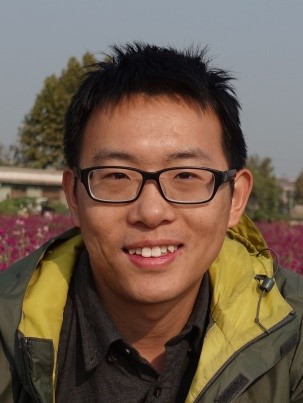}}]{Bo Peng}{\space}received B.Eng. degree in Beihang University and PhD degree in the Institute of Automation, Chinese Academy of Sciences in 2013 and 2018, respectively. Since July 2018, Dr. Bo Peng has joined the Institute of Automation, Chinese Academy of Sciences where he is currently an Associate Professor. His current research focuses on computer vision and image forensics. 

\end{IEEEbiography}

\begin{IEEEbiography}[{\includegraphics[width=1in,height=1.25in,clip,keepaspectratio]{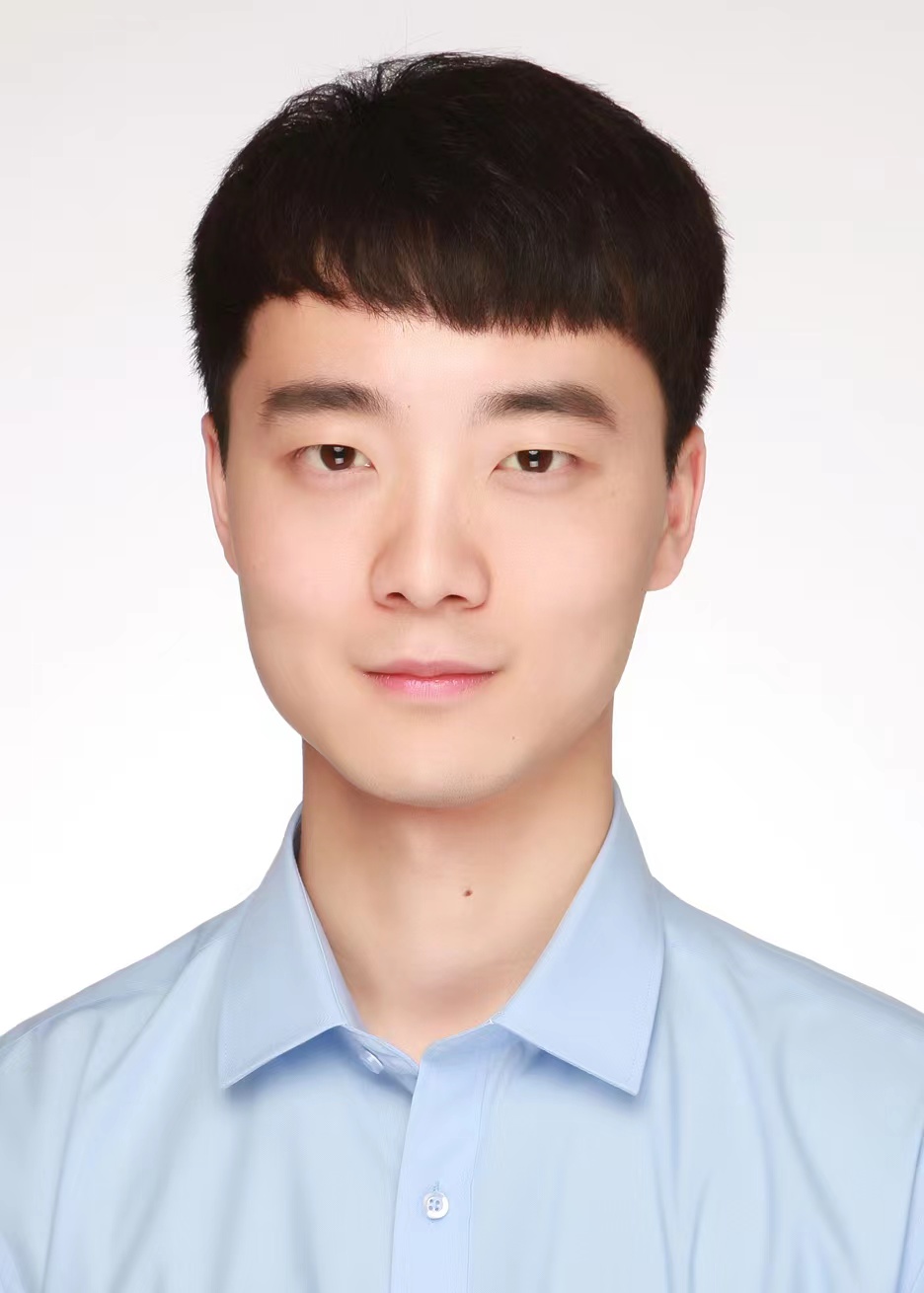}}]{Xu Wang}{\space}received B.Eng. Degree in Beijing Jiaotong University and Master Degree in Beihang University in 2014 and 2017, respectively. Since 2018, Xu Wang worked at ByteDance AI Lab and Intelligent Creation Team as computer vision algorithm researcher. His current research focuses on vision language model, text-to-image generation.

\end{IEEEbiography}

\begin{IEEEbiography}[{\includegraphics[width=1in,height=1.25in,clip,keepaspectratio]{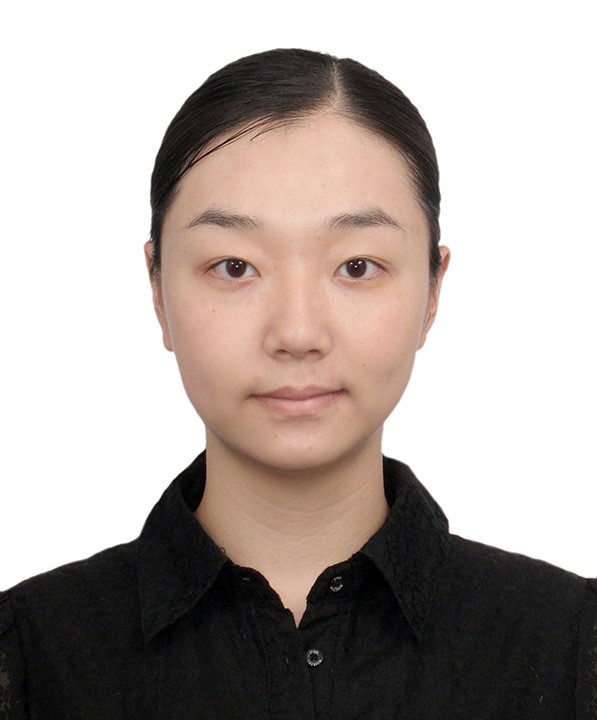}}]{Jing Dong}{\space}recieved her Ph.D in Pattern Recognition from the Institute of Automation, Chinese Academy of Sciences, China in 2010. Then she joined the Institute of Automation, Chinese Academy of Sciences and she is currently a Professor. Her research interests are towards Pattern Recognition, Image Processing and Digital Image Forensics including digital watermarking, steganalysis and tampering detection. She is a senior member of IEEE. She also has served as the deputy general of Chinese Association for Artificial Intelligence.

\end{IEEEbiography}
\vfill

\end{document}